\documentclass{article}

\usepackage{arxiv}

\usepackage[utf8]{inputenc} % allow utf-8 input
\usepackage[T1]{fontenc}    % use 8-bit T1 fonts
\usepackage{hyperref}       % hyperlinks
\usepackage{url}            % simple URL typesetting
\usepackage{booktabs}       % professional-quality tables
\usepackage{amsfonts}       % blackboard math symbols
\usepackage{nicefrac}       % compact symbols for 1/2, etc.
\usepackage{microtype}      % microtypography
\usepackage{lipsum}

\usepackage{varioref}
\usepackage{placeins}
\usepackage[Symbolsmallscale]{upgreek}
\usepackage{multirow}
\usepackage{siunitx}
\usepackage{mathtools}
\DeclarePairedDelimiter{\ceil}{\lceil}{\rceil}
\usepackage{arxiv}
\usepackage[utf8]{inputenc}
\usepackage[T1]{fontenc}
\usepackage{hyperref}
\usepackage{url}
\usepackage{booktabs}
\usepackage{amsfonts}
\usepackage{nicefrac}
\usepackage{microtype}
\usepackage{amsmath}
\usepackage{amssymb}
\usepackage{graphicx}
\usepackage{subcaption}
\usepackage{array}
\newcolumntype{P}[1]{>{\centering\arraybackslash}p{#1}}
\usepackage{textcomp}
\usepackage[table]{xcolor}
\usepackage{float}
\usepackage{enumitem}
\usepackage{comment}
\usepackage{soul}
\usepackage{color}
\usepackage[linesnumbered,vlined,titlenumbered,ruled]{algorithm2e}
\usepackage{xspace}

\newcommand{\fgrvref}[1]{Fig.~\vref{#1}}
\newcommand{\mathdot}{.}
\newcommand{\mathcomma}{,}

\newcommand{\betax}{\vecbeta^{\mbox{\scriptsize T}} \vec{x}_i}
\newcommand{\lbl}{l}
\newcommand{\vecbeta}{\boldsymbol{\upbeta}}
\newcommand{\secref}[1]{Section \ref{#1}}
\newcommand{\tabemph}[1]{\textbf{#1}}
\newcommand{\tabref}[1]{Table \ref{#1}}

\renewcommand{\vec}[1]{\mathbf{#1}}
\newcommand{\norm}[1]{\left\lVert#1\right\rVert}
\newcommand{\mat}[1]{\mathbf{#1}}
\newcommand{\randdropoff}{RandDropoff\xspace}
\newcommand{\cgp}{CGP\xspace}
\newcommand{\ncgpa}{NCGP-A\xspace}
\newcommand{\ncgpu}{NCGP\xspace}
\newcommand{\y}[2]{y^{#1}_{#2}}
\newcommand{\fgrref}[1]{Fig.~\ref{#1}}

\title{Estimating Latent Demand of Shared Mobility through Censored Gaussian Processes}

\author{
  Daniele ~Gammelli \\
  Danmarks Tekniske Universitet (DTU)\\
  Kgs. Lyngby, Denmark, 2800 \\
  \texttt{daga@dtu.dk} \\
  \And
  Inon ~Peled \\
  Danmarks Tekniske Universitet (DTU)\\
  Kgs. Lyngby, Denmark, 2800 \\
  \texttt{inonpe@dtu.dk} \\
  \And
  Filipe ~Rodrigues \\
  Danmarks Tekniske Universitet (DTU)\\
  Kgs. Lyngby, Denmark, 2800 \\
  \texttt{rodr@dtu.dk} \\
   \And
  Dario ~Pacino \\
  Danmarks Tekniske Universitet (DTU)\\
  Kgs. Lyngby, Denmark, 2800 \\
  \texttt{darpa@dtu.dk} \\
   \And
  Haci A.~Kurtaran \\
  Head of BI and Finances, Donkey Republic\\
  Copenhagen, Denmark \\
  \texttt{a\_kurtaran@hotmail.com} \\
   \And
  Francisco C.~Pereira \\
  Danmarks Tekniske Universitet (DTU)\\
  Kgs. Lyngby, Denmark, 2800 \\
  \texttt{camara@dtu.dk} \\
}

\begin{document}
\maketitle

\begin{abstract}
Transport demand is highly dependent on supply, especially for shared transport services where availability is often limited.
As observed demand cannot be higher than available supply, historical transport data typically represents a biased, or \emph{censored}, version of the \emph{true} underlying demand pattern.
Without explicitly accounting for this inherent distinction, predictive models of demand would necessarily represent a biased version of true demand, thus less effectively predicting the needs of service users.
To counter this problem, we propose a general method for censorship-aware demand modeling, for which we devise a \emph{censored} likelihood function. 
We apply this method to the task of shared mobility demand prediction by incorporating the censored likelihood within a Gaussian Process model, which can flexibly approximate arbitrary functional forms.
Experiments on artificial and real-world datasets show how taking into account the limiting effect of supply on demand is essential in the process of obtaining an unbiased predictive model of user demand behavior.
\end{abstract}


\section{Introduction}
\label{sec:introduction}
Being able to understand and predict demand is essential in the planning and decision-making processes of any given transport service, allowing service providers to make decisions coherently with user behavior and needs. 
Having reliable models of demand is especially relevant in shared transport modes, such as car-sharing and bike-sharing, where the high volatility of demand and the flexibility of supply modalities (e.g., infinitely many possible collocations of a car-sharing fleet) require that decisions be made in strong accordance with user needs.
If, for instance, we consider the bike sharing scenario, service providers face a great variety of complex decisions on how to satisfy user demand.
To name a few, concrete choices must be made for what concerns capacity positioning (i.e., where to deploy the service), capacity planning (i.e., dimensioning the fleet size), rebalancing (i.e., where and when to reallocate idle supply), and expansion planning (i.e., if and how to expand the reach of the service).
%Service providers thus face a great variety of complex decisions on how to satisfy user demand, including such matters as: capacity positioning (i.e., where to deploy the service), capacity planning (i.e., decide on supply size), rebalancing (i.e., where and when to reallocate idle supply), and expansion planning (i.e., if and how to expand the reach of the service).

Demand modeling uses statistical methods to capture user demand behavior based on recorded historical data. 
However, historical transport service data is usually highly dependent on \emph{historical supply} offered by the provider itself. 
In particular, supply represents an upper limit on our ability to observe realizations of the true demand.
For example, if we have data about a bike-sharing service with $100$ bikes available, we might observe a usage (i.e., demand) of $100$ bikes even though the actual demand might have potentially been higher.
This leads to a situation in which historical data is in fact representing a biased, or \emph{censored}, version of the underlying demand pattern in which we are truly interested.
More importantly, using censored data to build demand models will, as a natural consequence, produce a biased estimate of demand and an inaccurate understanding of user needs, which will ultimately result in non-optimal operational decisions for the service provider.

%Practically then, supply imposes an upper limit on our ability to observe realizations of the true demand.
%For example, if we have data about a bike-sharing service with $100$ bikes available, we can only observe usage rates (i.e., demand) as high as $100$ bikes simultaneously, even though the actual demand for bikes could sometimes be higher than $100$.
%This leads to a situation in which historical data is in fact representing a biased, or \emph{censored}, version of the underlying demand pattern in which we are truly interested. 
%Consequently, ignoring this inherent censorship while modeling historical data will by nature produce a biased estimate of demand and an inaccurate understanding of user needs, which in turn will result in sub-optimal operational decisions for service providers.

To address these problems, we propose a general approach for building models that are aware of the supply-censoring issue, and which are ultimately more reliable in reflecting user behavior.
To this end, we formulate a \emph{censored} likelihood function and apply it within a Gaussian Process model of user demand. 
Using both synthetic and real-world datasets of shared mobility services, we pit this model against non-censored models and analyze the conditions under which it is better capable of recovering true demand.

\section{Review of Related Work} \label{sec:related}

\subsection{Demand Forecasting for Shared Mobility}

There exists a long line of literature about demand forecasting, with research mainly focusing on obtaining predictions of demand based on historical data in combination with some exogenous input, such as weather and temporal information. 
The bike-sharing context alone already provides many such examples, as follows.
\cite{Yang, DU201939} use Random Forest models for rental predictions, whereas
\cite{rudloff} approach the demand forecasting problem through the use of Generalized Linear Models (GLMs) of counts, namely, Poisson and Negative-Binomial regression models. 
\cite{rixey} apply multivariate regression using data gathered from multiple bike-sharing systems.

\cite{Li} define a hierarchical prediction model, based on Gradient Boosting Regression Trees, for both rentals and returns at a city-aggregation-level.
They model rent proportions for clusters of bike-stations through a multi-similarity based inference and predict returns consistently with rentals by learning transition probabilities (i.e., where bikes end after being rented).
A different approach is found in \cite{Singhvi2015PredictingBU}, where the proposed prediction model is based on log-log regression and refers to the Origin-Destination (O-D) trip demand, rather than rentals and returns in specific locations or areas.
\cite{zhang2016deep, XU201847, Lin2018258} instead propose deep learning-based approaches to model bike in-flow and out-flow for the bike-sharing system areas and station-level demand respectively.

\subsection{Censored Modeling} \label{sec:rev_censored}

\emph{Data censorship} involves two corresponding sets of values: true and observed, so that each observed value may be \emph{clipped} above or below the respective true value.
Observations that have not been clipped are called \emph{non-censored}, and correspond exactly with true values.
All other observations are called \emph{censored} as they have been clipped at their observed values, therefore giving only a limited account of the corresponding true values (e.g., observed demand not corresponding with true demand).
We next review methods for modeling censored data, namely, \emph{censored modeling}.

An early form of censored modeling is the Probit model \cite{aldrich1984linear} for binary ($0$/$1$) observations, which assumes that the probability of observing $1$ depends linearly on the given explanatory features.
James Tobin extended this to a model for censored regression \cite{tobin}, now called Tobit, which assumes that the latent variable depends on the explanatory features linearly with Gaussian noise, and that observations are censored according to a fixed and known threshold. 
Because the censored observations divert Least Squares estimators away from the true, latent values, Tobit linear coefficients are instead commonly estimated via Maximum Likelihood techniques, such as Expectation Maximization or Heckman's Two-Step estimator \cite{amemiya1984tobit}.

The Tobit model has since become a cornerstone of censored modeling in multiple research fields, including econometrics and survival analysis \cite{Bewick2004}.
It has been extended through multiple variations \cite{Greene2011}, such as: multiple latent variables, e.g., one variable determines which observations are non-censored while another determines their values \cite{amemiya1985advanced}; Tobit-like models of censored count data \cite{terza1985tobit}; Tobit Quantile Regression \cite{powell1986censored}; dynamic, autoregressive Tobit \cite{wei1999bayesian}; and combination with Kalman Filter \cite{allik2015tobit}.
Other methods of censored regression have also been suggested, predominantly for survival analysis, such as: Proportional Hazard models \cite{kay1977proportional}, Accelerated Failure Time models \cite{wei1992accelerated}; Regularized Linear Regression \cite{li2016regularized}; and Deep Neural Networks \cite{biganzoli2002general, wu2018deep}.

As reflected in all the above works, research into censored modeling commonly aims to reconstruct the latent process that yields the true values.
In this work too, we aim to predict true values, not the occurrence of clipping nor the actually observed values.
Similarly to Tobit, we too assume that all observations are independent and that each observation is known to be either censored or non-censored.
However, contrary to all the above works, our method employs Gaussian Processes, which are non-parametric and do not assume any specific functional relationship between the explanatory features and the latent variable.
A theoretical treatment of non-parametric censored regression appears in \cite{lewbel2002nonparametric}, where the authors derive estimators for the latent variable with fixed censoring, analyze them mathematically, and apply them to a single, simulated dataset.
In contrast, this work proposes a likelihood function and applies it to multiple, real-world datasets with dynamic censoring.

\subsection{Common Approaches to Handling Demand Censorship}

Within the above stream of research, the censoring problem discussed in Section \ref{sec:introduction} is widely accepted.
However, to the best of our knowledge, there has been no extensive study on how observed demand can differ from the true, underlying demand and how these differences could impact predictive models. 
To assess this issue, common approaches regard various \emph{data cleaning techniques}.
For example, \cite{rudloff, omahony2015citibike, ALBINSKI201859, goh} attempt to avoid the bias induced by censored observations by filtering out the time periods where censoring might have occurred, before modeling.
As a further example, \cite{ALBINSKI201859}  substitute the censored observations with the mean of the historical (non-censored) observations regarding the same period.
A different but related approach is proposed by \cite{jian, Freund2019} who focus on obtaining an unbiased estimate of arrival rate by omitting historical instances where bikes were unavailable.

These common approaches manage, to some degree, to correct the bias characterizing observed demand. 
However, they also give relevance to the fact that this problem represents an important gap which requires further study in order to obtain reliable predictions for shared transport and other fields.
We believe there are two main reasons to why there is the need for a more structured view on the censoring problem:
(i) the approaches described above might not be applicable in many real-world scenarios in which the number of censored observations is very high, leading either to an excessive elimination of useful data or to an inadequate imputation of the censored data; (ii) rather than resorting to cleaning and imputation procedures as the ones above, it would be desirable to equip the forecasting model with some sort of awareness towards the censoring problem, so that the model can utilize the entire information captured in the observations to coherently adjust its predictions.

\section{Methodology}
\label{sec:methodology}

In this Section, we incrementally describe the building blocks of our proposed censored models.
First, we introduce several general concepts: Tobit likelihood, Gaussian Processes, and the role of \emph{kernels}.
Then, we combine these concepts by defining Censored Gaussian Processes along with a corresponding inference procedure.
% In what follows, $\mathcal{U}$ and $\mathcal{N}$ denote, respectively, the Uniform and Gaussian distributions, bold capitals denote matrices (e.g., $\mat{X}$, $\mat{K}$), and all other bold symbols denote column vectors (e.g., $\vec{y} = \left[y_1, \dots, y_n\right]^{\mbox{\scriptsize T}}$, $\vecbeta$, $\vec{0}$).

\subsection{Tobit Likelihood}

As a reference point for developing our censored likelihood function, let us now elaborate on the likelihood function of the popular Tobit censored regression model, described in \secref{sec:rev_censored}.
For each observation $y_i$, let $y_i^*$ be the corresponding true value.
For instance, in a shared transport demand setting, $y_i^*$ is the true, latent demand for shared mobility, while $y_i$ is the observed demand; if $y_i$ is non-censored then $y_i = y_i^*$, otherwise $y_i$ is censored so that $y_i < y_i^*$.
We are also given binary censorship labels $\lbl_i$, so that $\lbl_i = 1$ if $y_i$ is censored and $\lbl_i = 0$ otherwise (e.g., labels could be recovered by comparing observed demand to available supply).

Tobit parameterizes the dependency of 
$y_i^*$ 
on explanatory features
$\vec{x}_i$ 
through a linear relationship with parameters 
$\vecbeta$
and noise term 
$\varepsilon_i$, where all $\varepsilon_i$ are independently and normally distributed with mean zero and variance $\sigma^2$, namely:
\begin{equation}
    y_i^*= \betax + \varepsilon_i, \hspace{10mm} \varepsilon_i \sim \mathcal{N}(0,\,\sigma^{2})
    \mathdot
\label{eq:ystar}
\end{equation}
There are multiple variations of the Tobit model depending on where and when censoring arises.
In this work, without loss of generality, we deal with upper censorship, also known as \emph{Type I}, where $y_i$ is upper-bounded by a given threshold $y_u$, so that:
\begin{equation}
y_i=\begin{cases}
  y_i^*, & \text{if $y_i^* < y_u$} \\
  y_u, & \text{if $y_i^* \geq y_u$}
\end{cases}
\mathdot
\label{eq:yi}
\end{equation}
The likelihood function in this case can be derived from Eqs. \ref{eq:ystar} and \ref{eq:yi}, as follows.
\begin{enumerate}
\item If $\lbl_i = 0$, then $y_i$ is non-censored and so its likelihood is:
\begin{equation}
     \frac{1}{\sigma} \phi\left(\frac{y_i-\betax}{\sigma}\right)\mathcomma
\end{equation}
where $\phi$ is the standard Gaussian probability density function.
\item Otherwise, i.e., if $\lbl_i = 1$, then $y_i$ is censored and so its likelihood is:
\begin{equation}
    1 - \Phi\left(\frac{y_i-\betax}{\sigma}\right)\mathcomma
\end{equation}
where $\Phi$ is the standard Gaussian cumulative density function.
\end{enumerate}
Because all observations are assumed to be independent, their joint likelihood is: 
\begin{equation}
    \prod_{i} 
    \Bigg \{
    \frac{1}{\sigma} \phi\left(\frac{y_i-\betax}{\sigma}\right) 
    \Bigg \}^{1 - \lbl_i} 
    \Bigg \{
    1 - \Phi\left(\frac{y_i-\betax}{\sigma}\right)
    \Bigg \}^{\lbl_i}
    \mathcomma
\label{eq:tobitlik}
\end{equation}
which is a function of $\vecbeta$ and $\sigma$.

\subsection{Gaussian Processes}
\label{sec:gaussian_processes}

Gaussian Processes (GPs) \cite{rasmussen2005gaussian} are an extremely powerful and flexible tool belonging to the field of probabilistic machine learning \cite{Ghahramani2015}. 
GPs have been applied successfully to both classification and regression tasks regarding various transport related scenarios such as travel times \cite{rodr_boris, Ide2009}, congestion models \cite{Liu:2013:ACR:2487575.2487598}, crowdsourced data \cite{rodr_pereira, rodr_Henri}, traffic volumes \cite{Xie}, etc. 
For example, given a finite set of points for regression, there are typically infinitely many functions which fit the data, and GPs offer an elegant approach to this problem by assigning a probability to every possible function. 
Moreover, GPs implicitly adopt a full probabilistic approach, thus enabling the structured quantification of the confidence -- or equivalently, the uncertainty -- in the predictions of a GP model.
This ease in uncertainty quantification is one of the principal reasons why we chose to use GPs for demand prediction in the shared mobility domain. 
Indeed, transport service providers are not only interested in more accurate demand models, but also, and maybe most importantly, wish to make operational decisions based on the measure with which the model is confident of its predictions.

Given a dataset $\mathcal{D} = \{(\vec{x}_i,y_i )\}_{i=1}^n$ with $n$ input vectors $\vec{x}_i$ and scalar outputs $y_i$, the goal of GPs is to learn the underlying data distribution by defining a distribution over functions.
GPs model this distribution by placing a multivariate Gaussian distribution defined by
a \emph{mean function} $m(\vec{x})$ and a \emph{covariance function} $k(\vec{x}, \vec{x}^{\prime})$ (between all possible pairwise combinations of input points $\{\vec{x}, \vec{x}^{\prime}\}$ in the dataset), or \emph{kernel}, on a latent variable $\vec{f}$. More concretely, a GP can be seen as a collection of random variables which have a joint Gaussian distribution. GPs are therefore a Bayesian approach which assumes a priori that function values behave according to:
\begin{equation}
    p(\vec{f} | \vec{x_1}, \dots, \vec{x_n}) = \mathcal{N}(\vec{m}, \mat{K})
    \mathcomma
\label{eq:f}
\end{equation}
where $\vec{f} = [f_1, \dots ,f_n]^{\mbox{\scriptsize T}}$ is the vector of latent random variables, $\vec{m}$ is a mean vector, and $\mat{K}$ is a covariance matrix with entries defined by a covariance function $k$, so that $\mat{K}_{i,j} = k(\vec{x}_i,\vec{x}_j)$.
As is customary in GP modeling, we shall assume without loss of generality that the joint Gaussian distribution is centered on $\vec{m} \equiv \vec{0}$.
Also, because the response variable is continuous, we model each $y_i$ as generated from a Gaussian distribution centered on $f_i$ with noise variance $\sigma^2$.

\subsection{Kernels}
\label{sec:kernels}
A fundamental step in Gaussian Process modeling is the choice of the covariance matrix $\mat{K}$.
This not only describes the shape of the underlying multivariate Gaussian distribution, but also determines the characteristics of the function we want to model.
Intuitively, because entry $\mat{K}_{i, j}$ defines the correlation between the $i$-th and the $j$-th random variables, the kernel describes a measure of similarity between points, ultimately controlling the shape that the GP function distribution adopts.

Multiple kernels can be combined, e.g., by addition and multipication, to generate more complex covariance funcitons, thus allowing for a great flexibility and better encoding of domain knowledge in the regression model.
For the datasets in our experiments (\secref{sec:experiments}), we use different combinations of the following three kernels, where $
\norm{\vec{x} - \vec{x}^\prime}$
denotes the Euclidean distance between vectors $\vec{x}, \vec{x}^\prime \in \mathbb{R}^n$:

\begin{enumerate}
    \item \emph{Squared Exponential Kernel (SE)}
    \begin{equation}
        k_{SE}(\vec{x}, \vec{x}^\prime) = \lambda^2 \exp{\left(-\frac{(\norm{\vec{x}-\vec{x}^\prime})^2}{2 \tau ^2}\right)}
    \label{eq:rbf}
    \end{equation}
    The SE kernel (also called RBF: Radial Basic Function) intuitively encodes the idea that nearby points should be correlated, therefore generating relatively smooth functions.
    The kernel is parameterized by variance $\lambda^2$ and length scale $\tau$: for larger $\tau$, farther points are considered more similar; $\lambda$ acts as an additional scaling factor, so that larger $\lambda$ corresponds to higher spread of functions around the mean.

    \item \emph{Periodic Kernel}
    \begin{equation}
        k_{Per}(\vec{x}, \vec{x}^\prime) = \lambda^2 \exp{\left(-\frac{2\sin^2(\pi \norm{\vec{x}-\vec{x}^\prime}/\rho)}{\tau^2}\right)}
    \label{eq:periodic}
    \end{equation}
    The Periodic kernel allows modeling functions with repeated patterns, and can thus extrapolate seasonality in time-series data.
    Parameters $\tau$ and $\lambda^2$ have the same effect as for the SE kernel, while parameter $\rho$ corresponds to period length of patterns.
    
    \item \emph{Mat\'ern Kernel}
    \begin{equation}
        k_{Mat}(\vec{x}, \vec{x}^\prime) = \lambda^2 \frac{2^{1-\nu}}{\Gamma(\nu)} \left(\sqrt{2\nu}\frac{\norm{\vec{x} - \vec{x}^\prime}}{\tau}\right)^{\nu} K_{\nu}\left(\sqrt{2\nu}\frac{\norm{\vec{x} - \vec{x}^\prime}}{\tau}\right)
        \mathcomma
    \label{eq:matern}
    \end{equation}
    where $\Gamma$ is the Gamma function, $K_{\nu}$ is the Bessel function \cite{mabramowitz64:handbook}, and as before, $\tau$ is lengthscale and $\lambda$ is variance. 
    The Mat\'ern kernel can be considered as a generalization of the SE kernel, parameterized by positive parameters $\nu$ and $\tau$, where lower values of $\nu$ yield less smooth functions.
\end{enumerate}

As is common, we select kernel parameters in our experiments through Type-II Maximum Likelihood, also known as Empirical Bayes, whereby latent variables are integrated out.

\subsection{Gaussian Process with Censored Likelihood} \label{sec:censored gaussian processes}
We have so far presented two separate modeling approaches: Tobit models for censored data, and non-parametric Gaussian Processes for flexible regression of arbitrarily complex patterns.
For non-parametric censored regression, we now combine the two approaches within one model by defining a censorship-aware likelihood function for GPs:
\begin{equation}
    \prod_{i} 
    \Bigg \{
    \frac{1}{\sigma} \phi\left(\frac{y_i - f_i}{\sigma}\right)
    \Bigg \}^{1 - \lbl_i}
    \Bigg \{
    1 - \Phi\left(\frac{y_i - f_i}{\sigma}\right)
    \Bigg \}^{\lbl_i}
    \mathdot
\label{eq:gplik}
\end{equation}
Eq. \ref{eq:gplik} is obtained from the Eq. \ref{eq:tobitlik} by replacing $\betax$, the Tobit prediction, with $f_i$, the GP prediction.
We next propose an inference procedure for obtaining the posterior distribution of Censored GP with this likelihood function.

\subsubsection{Inference}
\label{inference}
In a Bayesian setting, we are interested in computing the posterior over the latent $\vec{f}$, given the response $\vec{y}$ and features $\mat{X}$.
Bayes' rule defines this posterior exactly:
\begin{equation}
    p(\vec{f}|\vec{y}, \mat{X}) = \frac{p(\vec{f})p(\vec{y}|\vec{f}, \mat{X})}{p(\vec{y}|\vec{X})}\mathdot
\label{eq:bayes}
\end{equation}
However, exact calculation of the denominator in Eq. \ref{eq:bayes} is intractable because of the Censored GP likelihood (Eq. \ref{eq:gplik}), as also explained in \cite{rasmussen2005gaussian}. 
The posterior distribution thus needs to be estimated approximately, which we do in this study via \emph{Expectation Propagation} (EP) \cite{minka2001expectation}.
% Common approaches to approximate bayesian inference regard variations of Markov Chain Monte Carlo sampling (MCMC) \cite{Hastings, Gelfand} and Variational Inference (VI) \cite{Blei_2017}. 
EP alleviates the denominator intractability by approximating the likelihood of each observation through an un-normalized Gaussian in the latent $f_i$, as:
\begin{equation}
    p(y_i|f_i) \simeq t_i(f_i|\Tilde{Z}_i, \Tilde{\mu}_i, \Tilde{\sigma}_i^2) \triangleq \Tilde{Z}_i \mathcal{N}(f_i|\Tilde{\mu}_i, \Tilde{\sigma}_i^2)
    \mathcomma
\label{eq:12}
\end{equation}
where $t_i$ is the $i$-th factor, or site, with site parameters $\Tilde{Z}_i, \Tilde{\mu}_i, \Tilde{\sigma}_i^2$.
The likelihood is then approximated as a product of the $n$ independent local likelihood approximations:
\begin{equation}
    \prod_{i=1}^{n}t_i(f_i|\Tilde{Z}_i, \Tilde{\mu}_i, \Tilde{\sigma}_i^2) = \mathcal{N}(\Tilde{\vec{\mu}}, \Tilde{\mat{\Sigma}})\prod_{i=1}^{n}\Tilde{Z}_i
    \mathcomma
\label{eq:13}
\end{equation}
where $\Tilde{\vec{\mu}} = \left[\Tilde{\mu}_1, \dots, \Tilde{\mu}_n\right]$, and $\Tilde{\mat{\Sigma}}$ is a diagonal covariance matrix with $\Tilde{\mat{\Sigma}}_{ii} = \Tilde{\sigma}_i^2$.
Consequently, the posterior $p(\vec{f}|\vec{y}, \mat{X})$ is approximated by
\begin{equation}
    q(\vec{f}|\vec{y}, \mat{X}) \triangleq \frac{1}{Z_{EP}}p(\vec{f}|\mat{X})\prod_{i=1}^{n}t_i(f_i|\Tilde{Z}_i, \Tilde{\mu}_i, \Tilde{\sigma}_i^2) = \mathcal{N}(\vec{\mu}, \mat{\Sigma})
    \mathcomma
\label{eq:14}
\end{equation}
where $\mat{\Sigma} = (\mat{K}^{-1} + \Tilde{\mat{\Sigma}}^{-1})^{-1}$, $\vec{\mu} = \mat{\Sigma}\Tilde{\mat{\Sigma}}^{-1}\Tilde{\vec{\mu}}$, and $Z_{EP}$ is the EP approximation to the marginal likelihood.

The key idea in EP is to update the single site approximation $t_i$ sequentially by iterating the following four steps.
Firstly, given a current approximation of the posterior, the current $t_i$ is left out from this approximation, defining a \emph{cavity distribution}:
\begin{equation}\label{eq:15}
\begin{split}
q_{-i}(f_i) & \propto \int p(\vec{f}|\mat{X})\prod_{j\neq i} t_i(f_i|\Tilde{Z}_i, \Tilde{\mu}_i, \Tilde{\sigma}_i^2) df_j
\mathcomma
\\
q_{-i}(f_i) & \triangleq \mathcal{N}(f_i|\mu_{-i}
\mathcomma
\sigma_{-i}^{2}),\\
\end{split}
\end{equation}
where $\mu_{-i} = \sigma_{-i}^{2}(\sigma_{i}^{-2}\mu_{i} - \Tilde{\sigma}_{i}^{-2}\Tilde{\mu}_{i})$ and $\sigma_{-i}^{2} = (\sigma_{i}^{-2} - \Tilde{\sigma}_{i}^{-2})^{-1}$.

The second step is the computation of the target non-Gaussian marginal combining the cavity distribution with the exact likelihood $p(y_i|f_i)$ :
\begin{equation}
    q_{-i}(f_i)p(y_i|f_i)
    \mathdot
\label{eq:16}
\end{equation}
Thirdly, a Gaussian approximation to the non-Gaussian marginal $\hat{q}(f_i)$ is chosen such that it best approximates the product of the cavity distribution and the exact likelihood:
\begin{equation}
    \hat{q}(f_i) \triangleq \hat{Z}_i \mathcal{N}(\hat{\mu}_i, \hat{\sigma}_i^{2}) \simeq q_{-i}(f_i)p(y_i|f_i)
    \mathdot
\label{eq:17}
\end{equation}
To achieve this, EP uses the property that if $\hat{q}(f_i)$ is Gaussian, then the distribution which minimizes the Kullback-Leibler divergence $\mbox{KL}\left(q_{-i}(f_i)p(y_i|f_i)\|\hat{q}(f_i)\right)$, is the distribution whose first and second moments match the ones of $q_{-i}(f_i)p(y_i|f_i)$.
In addition, given that $\hat{q}(f_i)$ is un-normalized, EP also imposes the condition that the zero-th moment should match. 
Formally then, EP minimises the following objective:
\begin{equation}
    \mbox{KL}\left(q_{-i}(f_i)p(y_i|f_i)\|\hat{q}(f_i)\right) = \mathbb{E}_{q_{-i}(f_i)p(y_i|f_i)}\left[ \log \left(\frac{q_{-i}(f_i)p(y_i|f_i)}{\hat{q}(f_i)}\right) \right].
\end{equation}
Lastly, EP chooses the $t_i$ for which the posterior has the marginal from the third step.

\subsubsection{Censored Likelihood Moments}
\label{censored likelihood moments}
The implementation of EP in our experiments is based on GPy \footnote{\url{https://sheffieldml.github.io/GPy/}}, an open source Gaussian Processes package in Python.
We have implemented the censored likelihood and defined the following moments for the EP inference procedure:
\begin{equation}\label{eq:18_1}
\begin{split}
\hat{Z}_i & = \int q_{-i}(f_i)t_i df_i
\mathcomma
\\
\hat{\mu}_i & = \mathbb{E}_q[f_i]
\mathcomma
\\
\hat{\sigma}_i & = \mathbb{E}_q[(f_i - \mathbb{E}_q[f_i])^2]
\mathdot
\end{split}
\end{equation}
The moments of our censored likelihood (Eq. \ref{eq:gplik}) can be defined analytically, depending on the censoring upper limit $y_u$, as follows (for a detailed derivation, see \cite{groot2012gaussian}).
\begin{itemize}
\item If $y_i$ < $y_u$, then:
\begin{equation}\label{eq:18_2}
\begin{split}
\hat{Z}_i & = \frac{1}{\sqrt{2\pi(\sigma^2 + \sigma_{-i}^2)}} \exp{\left(- \frac{(y_i - \mu_{-i})^2}{2(\sigma^2 + \sigma_{-i}^2)}\right)}
\mathcomma
\\
\hat{\mu}_i & = \mu_{-i} + \sigma_{-i}^2\left(\frac{y_i-\mu_{-i}}{\sigma^2 + \sigma_{-i}^2}\right)
\mathcomma
\\
\hat{\sigma}_i & = \sigma_{-i}^2 -  \sigma_{-i}^4\left(\frac{1}{(\sigma^2 + \sigma_{-i}^2)}\right)
\mathdot
\end{split}
\end{equation}

\item Otherwise, $y_i$ = $y_u$, and:
\begin{equation}\label{eq:18_3}
\begin{split}
\hat{Z}_i & = \Phi(\bar{z}_i)
\mathcomma
\\
\hat{\mu}_i & = \mu_{-i} + \frac{\sigma_{-i}^2\mathcal{N}(\bar{z}_i)}{\Phi(\bar{z}_i)\sqrt{\sigma^2 + \sigma_{-i}^2}}
\mathcomma
\\
\hat{\sigma}_i & = \sigma_{-i}^2 - \frac{\sigma_{-i}^4\mathcal{N}(\bar{z}_i)}{\Phi(\bar{z}_i)(\sigma^2 + \sigma_{-i}^2)}\left(\bar{z}_i + \frac{\mathcal{N}(\bar{z}_i)}{\Phi(\bar{z}_i)}\right)
\mathcomma
\end{split}
\end{equation}
where $\Phi$ is the Gaussian Cumulative Density Function and $\bar{z}_i = \left(\mu_{-i} - y_u\right) \Big/ \sqrt{\sigma^2 + \sigma_{-i}^2}$.
\end{itemize}

\section{Experiments} \label{sec:experiments}

In this Section, we execute experiments for several datasets as follows.
First, to convey a high-level intuition on how censored models can have an advantage compared to non-censored models, we start with two relatively simple datasets: synthetically generated data in Section \ref{sec:synthetic}, and real-world motorcycle accident data in Section \ref{sec:motorcycle}.
Equipped with this intuition, we then proceed to more elaborate real-world datasets.
In Section \ref{sec:donkey}, we then build models to predict bike pickups for a major bike-sharing service provider in Copenhagen, Denmark.
Then in Section \ref{sec:taxi}, we use data about yellow taxis in New York City.
For both bike-sharing and taxis, pickups thus represent mobility demand, which is affected by the censoring phenomenon introduced in previous sections because of finite supply.

\subsection{Models} \label{sec:models}
As introduced in previous sections, the focus of our experiments\footnote{Source code available at: \url{https://github.com/DanieleGammelli/CensoredGP}} is the comparison between \emph{Censored} and \emph{Non-Censored} models in the estimation of true demand patterns. 
We thus compare three GP models:
\begin{enumerate}[label=(\roman*)]
    \item \emph{Non-Censored Gaussian Process (\ncgpu)}: represents the Gaussian Process model most commonly used in literature, i.e., with Gaussian observation likelihood. \ncgpu is trained on the entire dataset, consisting of both censored and non-censored observations, without discerning between them.
    
    \item \emph{Non-Censored Gaussian Process, Aware of censorship (\ncgpa)}: functionally equivalent to \ncgpu, but uses information on censoring as a pre-processing step.
    That is, \ncgpa is trained only on non-censored points, thus avoiding exposure to a biased version of the true demand (because of censoring). This, however, comes at the cost of ignoring relevant information embedded in the censored data
    
    \item \emph{Censored Gaussian Process (\cgp)}: this model considers all observations -- censored and non-censored -- through the likelihood function defined in Eq. \ref{eq:gplik} (Section \ref{sec:censored gaussian processes}).
\end{enumerate}

\subsection{Synthetic Dataset} \label{sec:synthetic}

\begin{figure}[tb]
\begin{subfigure}{0.49\textwidth}
\includegraphics[width=\linewidth]{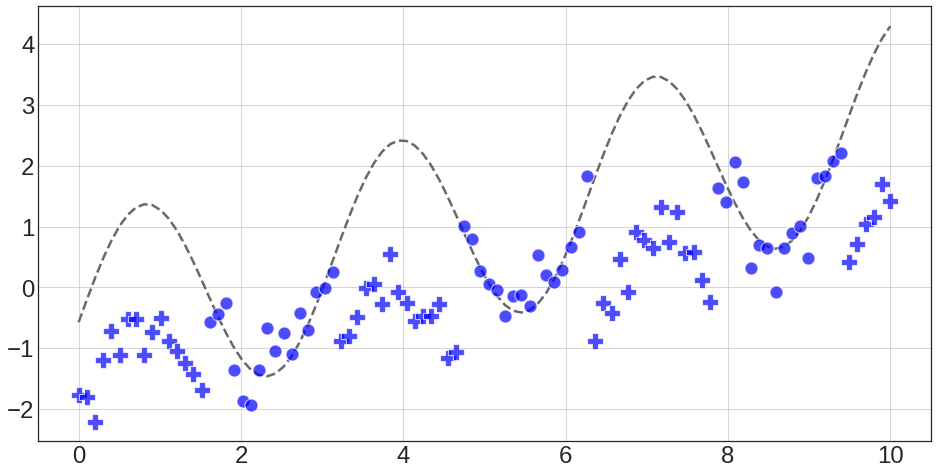}
\caption{Synthetic Data} \label{fig:1a}
\end{subfigure}
\hspace*{\fill} % separation between the subfigures
\begin{subfigure}{0.49\textwidth}
\includegraphics[width=\linewidth]{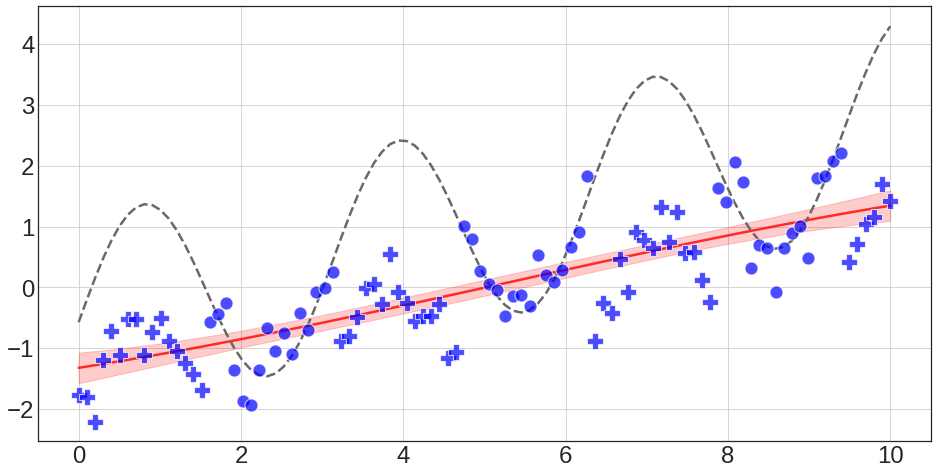}
\caption{\ncgpu} \label{fig:1b}
\end{subfigure}
\vspace*{\fill} % separation between the subfigures
\begin{subfigure}{0.49\textwidth}
\includegraphics[width=\linewidth]{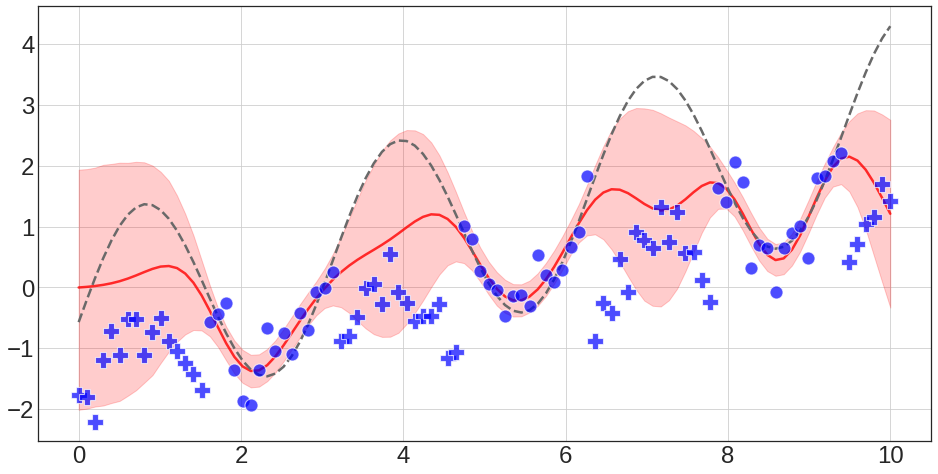}
\caption{\ncgpa} \label{fig:1c}
\end{subfigure}
\hspace*{\fill} % separation between the subfigures
\begin{subfigure}{0.49\textwidth}
\includegraphics[width=\linewidth]{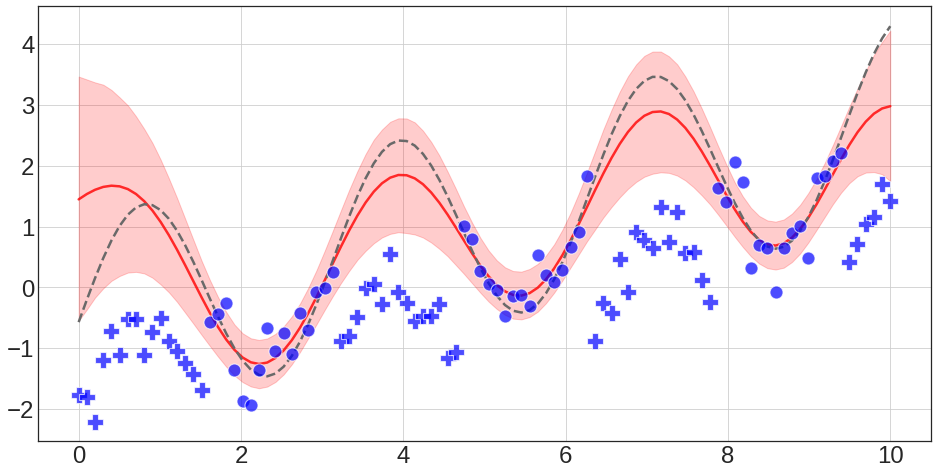}
\caption{\cgp} \label{fig:1d}
\end{subfigure}
\caption{Synthetically generated data (a) and model fits for \ncgpu (b), \ncgpa (c) and \cgp (d). 
The dashed line is the latent true function $f^*$;  crosses and circles are, respectively, censored and non-censored observations; the continuous line and the shaded area are, respectively, the posterior mean over the GP latent variable $\vec{f}$ and the corresponding posterior $95\%$ confidence interval.}
\label{fig:1}
\end{figure}

To demonstrate the capabilities of our approach, we begin with a simple, synthetically generated dataset, where the censoring pattern has a relevant and meaningful structure.
We define a latent function for $x \in \mathbb{R}$:
\begin{equation}
f^*(x) = 2 + \frac{\sin{(2x)}}{2} + \frac{x}{10}
\mathcomma
\end{equation}
which we wish to recover from censored, noisy observations. 
We generate these observations for  $x=\left[0, 10\right]$ by repeatedly evaluating $f^*$ and adding a noise term, which is independently sampled from $\mathcal{N}(0, \sigma^2)$ with $\sigma^2 = 0.1$. 
As seen in \fgrref{fig:1a}, this censoring process generates a cloud of points far from the true dynamics of the latent $f^*$ in the peaks of the sinusoidal oscillation.

We fit \ncgpu, \ncgpa and \cgp to the observations and measure how well each of them recovers the true $f^*$. 
The resulting predictions by the three models in \fgrref{fig:1b}, \ref{fig:1c}, \ref{fig:1d} highlight interesting aspects worth underlying:
\begin{itemize}
    \item \fgrref{fig:1b}: \ncgpu predicts an approximately linear trend for observed data, which, if we did not know the true function behind our observations, would seem like a reasonable conclusion. Nevertheless, the predictions are definitely far from the true data generating process.
    \item \fgrref{fig:1c}: By discarding the censored observations, \ncgpa is able to correctly assess the behavior of the underlying function in regions close to observable data. However, \ncgpa is not able to generalize to regions affected by censoring.
    \item \fgrref{fig:1d}: \cgp, on the other hand is able to exploit its knowledge of censoring to make better sense of observable data. 
    Through the use of a censored likelihood (Eq. \ref{eq:gplik}), \cgp assigns higher probability to plausible functional patterns, which are coherent not only with the observable data but also with the censored function.
\end{itemize}

\subsection{Real-World Dataset 1: Motorcycle Accident} \label{sec:motorcycle}

\begin{figure}[!t]
    \centering
    \includegraphics[width=\textwidth]{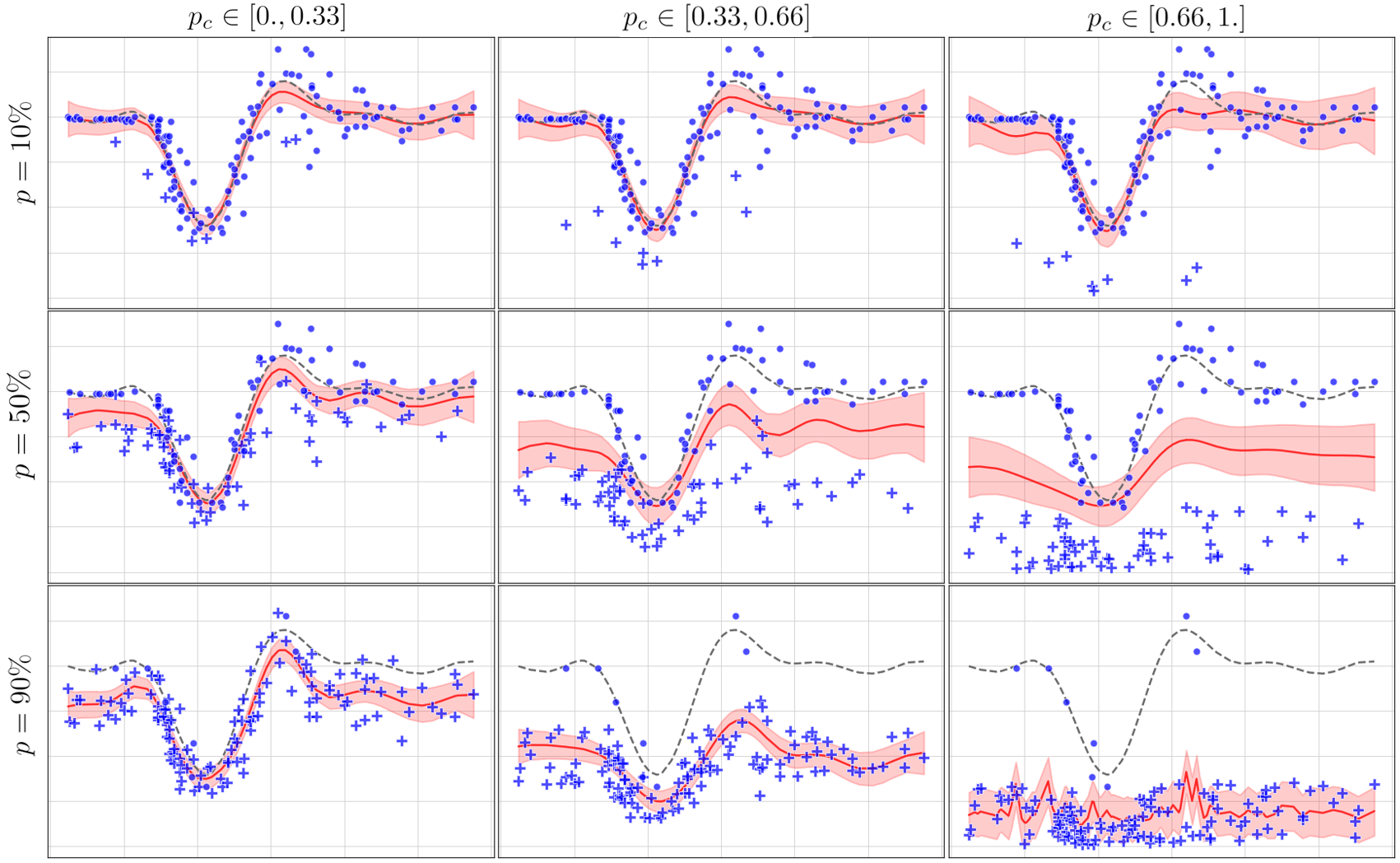}
    \caption{\ncgpu fit for intensifying censorship of the motorcycle accident data.}
    \label{fig:motorcycle_ncgpu}
\end{figure}

\begin{figure}[!t]
    \centering
    \includegraphics[width=\textwidth]{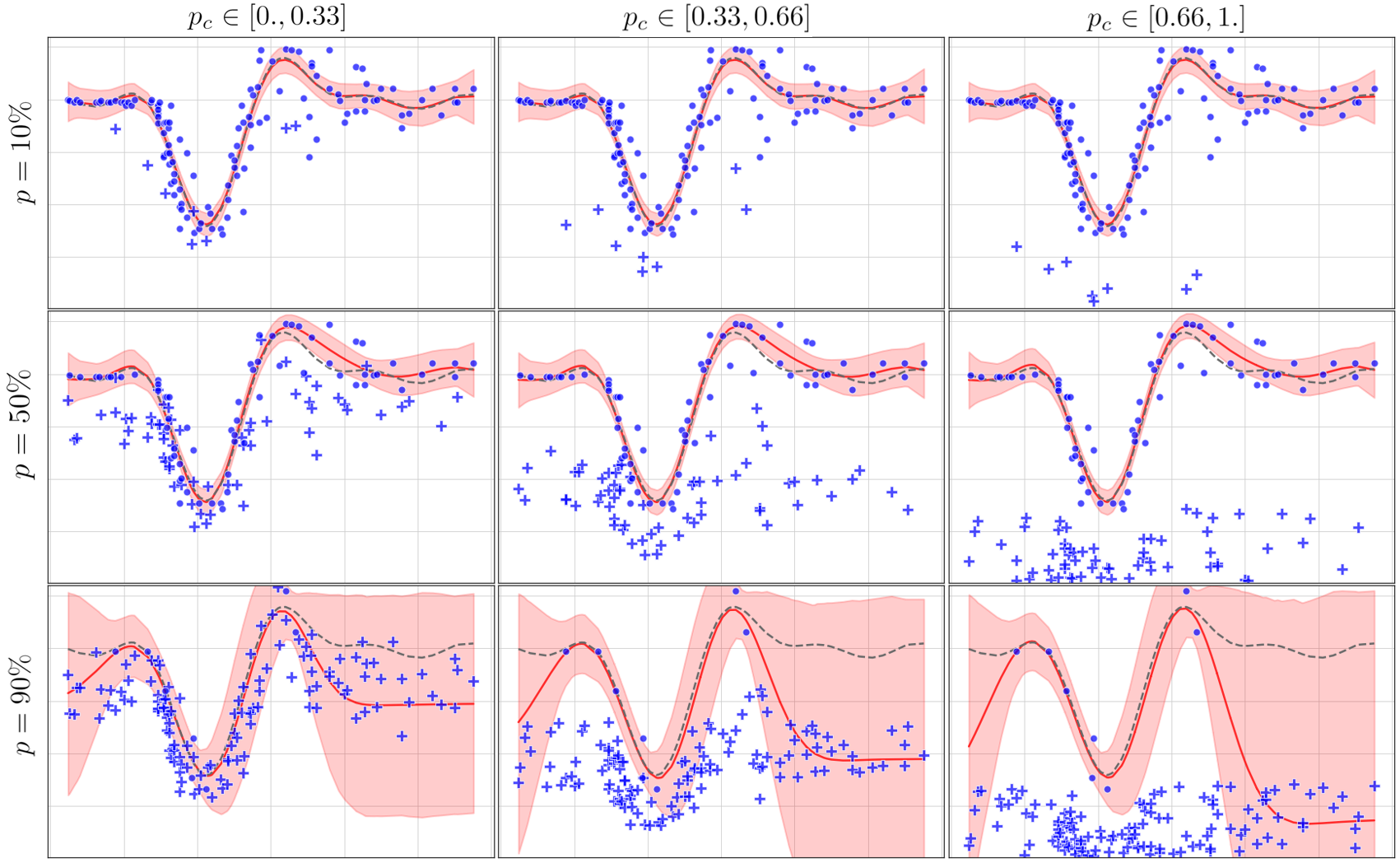}
    \caption{\ncgpa fit for intensifying censorship of the motorcycle accident data.}
    \label{fig:motorcycle_ncgpa}
\end{figure}

\begin{figure}[!t]
    \centering
    \includegraphics[width=\textwidth]{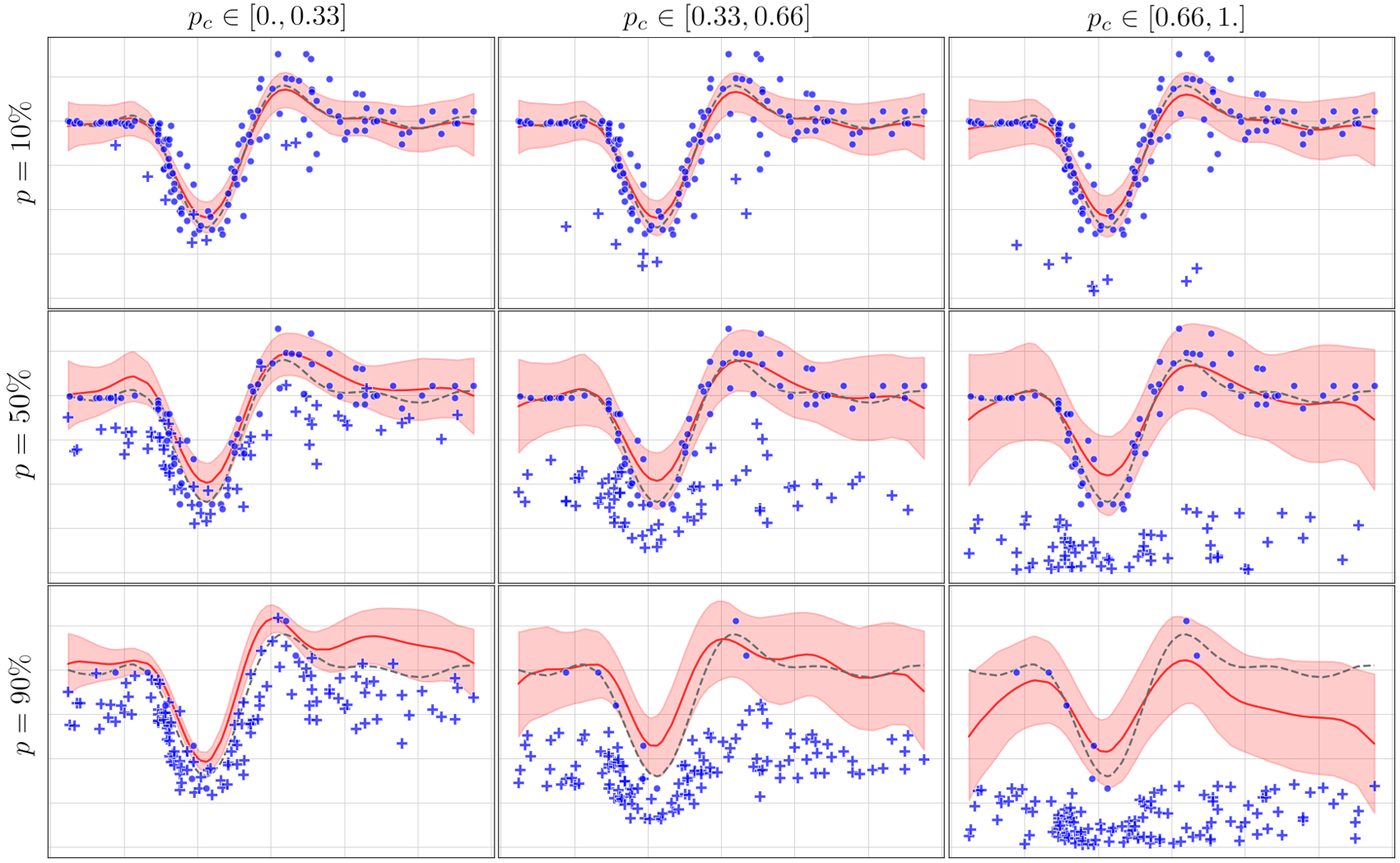}
    \caption{\cgp fit for intensifying censorship of the motorcycle accident data.}
    \label{fig:motorcycle_cgp}
\end{figure}

Having demonstrated our approach on an artificial dataset, we now make use of the openly accessible Motorcycle Accident dataset \cite{motorcycle} used in literature for a wide variety of statistical learning tasks.
The data consists of acceleration measurements over $133$ consecutive $\si{\ms}$.
For this experiment, we aim at exploring how variations in the severity of the censoring process can impact the models' performance.
In particular, we will be focusing on two axes of variation: the \emph{percentage} of censored points, which controls how often the true function is observable, and the \emph{intensity} of censoring, which controls how far censored observations are from true values. 
Concretely, we apply the following censoring process in the experiments with Motorcycle data $y^*_1, \dots, y^*_n$, given any $0 \leq p \leq 1 $ and $0 \leq a < b \leq 1$.
\begin{enumerate}
    \item Initialize $y_i = y^*_i$ for all $i=1..n$.
    \item Define the set of censored observations $N_c$ by randomly selecting $\ceil{pn}$ elements from $1..n$.
    \item For each $i \in N_c$, independently sample a censorship intensity as $p_c^{(i)} \sim \mathcal{U}\left[a, b\right]$, and censor the $i$'th observation as $y_i = \left( 1-p_c^{(i)} \right) y_i^*$.
\end{enumerate}

We explore moderate to extreme censoring through increasing percentage of censored points, as $p=0.1, 0.5, 0.9$, and increasing censorship intensity, as $\left[a, b\right] = \left[0, 0.33\right], \left[0.33, 0.66\right], \left[0.66, 1.0\right]$.
Figs. \ref{fig:motorcycle_ncgpu}, \ref{fig:motorcycle_ncgpa} and \ref{fig:motorcycle_cgp} show how each of \ncgpu, \ncgpa and \cgp , respectively, reacts to these variations in censorship intensity:
the percentage of censored observations increases from top to bottom, and censorship intensity increases from left to right.
In each plot, the horizontal and vertical axes correspond to time and acceleration, respectively; we omit axis labels to reduce visual clutter, as also the specifics of this dataset are less important in the context of this experiment.
We see that:
\begin{itemize}
    \item \fgrref{fig:motorcycle_ncgpu}: Despite being able to successfully recover the latent function for small censorship intensities (first row), \ncgpu is then exposed to an excessive amount of bias because of the increased censoring.
    This results in completely distorted predictions, as the remaining non-censored points are essentially considered as outliers by the model. 
    \item \fgrref{fig:motorcycle_ncgpa}: On the other hand, \ncgpa manages to avoid the bias coming from censored points by ignoring them, as long as the non-censored observations characterize well the latent function. 
    When this condition no longer holds, \ncgpa is unable to generalize to unobserved domains, so that it yields biased predictions with high uncertainty.
    \item \fgrref{fig:motorcycle_cgp}: Lastly, \cgp not only manages to avoid the bias introduced by censored observations in moderate to high censoring scenarios, but is also able to couple information coming from both censored and non censored observations to achieve better predictions in regions of extreme censoring.
\end{itemize}

\subsection{Real-World Dataset 2: Bike-Sharing Demand and Supply} \label{sec:donkey}
In this Section, we deal with the problem of building a demand prediction model for a bike-sharing system.
We use real-world data provided by Donkey Republic, a major bike-sharing provider in the city of Copenhagen, Denmark. 
Donkey Republic can be considered a hub-based service, meaning that the user of the service is not free to pick up or drop off a bike in any location, but is restricted to a certain number of virtual hubs around Copenhagen. 
Our objective is to model daily rental demand in the hub network. 

\subsubsection{Data Aggregation}

\begin{figure}[tb]
    \centering
    \includegraphics[width=0.7
    \linewidth]{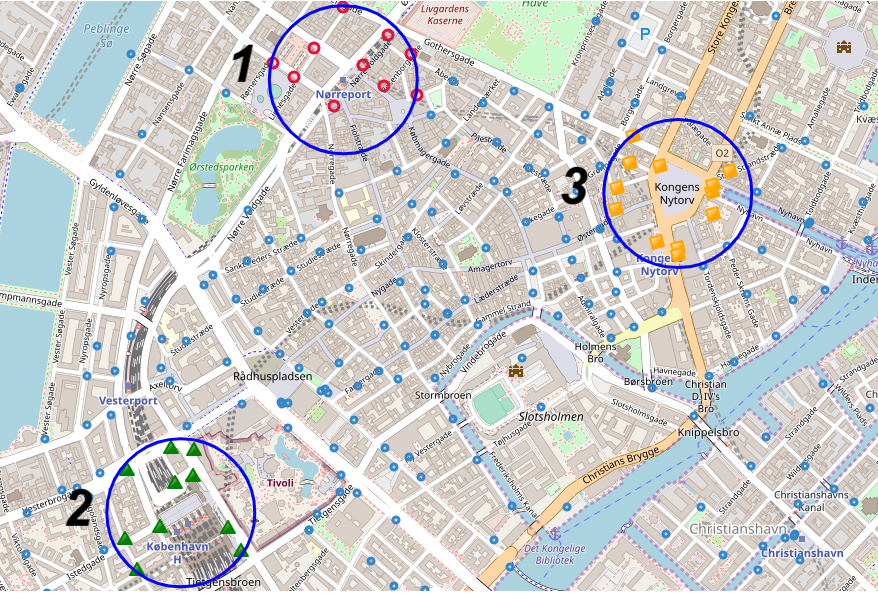}
    \caption{Map with: 1) marked locations of Donkey Republic hubs, 2) the three super-hubs in our experiments, as big circles around constituent hubs.}
    \label{fig:superhubs}
\end{figure}

The given data consists of individual records of users renting and returning bikes in hubs during $379$ days: from 1 March 2018 until 14 March 2019.
Hence before modeling daily rentals, we aggregate the data both spatially and temporally.
Spatially, $32$ hubs were aggregated in three \emph{super-hubs} by selecting three main service areas (such as the central station and main tourist attractions) and considering a $\SI{200}{\meter}$ radius around these points of interest (\fgrref{fig:superhubs}).
The choice of radius is justified by Donkey Republic business insights, whereby users typically agree to walk at most $\SI{200}{\meter}$ to rent a bike and typically give up if the bike is farther.
Temporally, the data at our disposal allowed us to retrieve the time-series of daily rental pickups regarding the three super-hubs, which will represent the target of our prediction model.

The spatial aggregation of individual hubs into super-hubs is an important modeling step in building the prediction model, for two main reasons:
\begin{enumerate}
    \item Time-series for individual hubs reveal excessive fluctuations in rental behavior. 
    Hence separate treatment of individual hubs could likely hide most of the valuable regularities in the data and ultimately expose the predictive models to an excessive amount of noise.
    \item As seen in \fgrref{fig:superhubs}, the individual hubs are very close one to the other, especially in central areas of the city. 
    Demand patterns between neighboring hubs can thus be well correlated, and a bike-sharing provider would actually benefit more from understanding the demand in an entire area rather than in a single hub.
    Moreover, bike-sharing demand conceptually covers an entire area, as bike-sharing users would likely walk a few dozen meters more to rent a bike.
    Hence from here on, we assume that if no bikes are available at some hub, then users who wish to rent a bike are willing to walk to any other hub in the same super-hub.
\end{enumerate}

\subsubsection{Censorship of Bike-Sharing Data} \label{sec:bikecens}

Ideally and before modeling, we would like to have access to the true bike-sharing demand, free of any real-world censorship.
However, this ideal setting is impossible, as historical data records are necessarily censored intrinsically to some extent.
Consequently and for the sake of experimentation, we assume that the given historical data represents true demand (which is what ideally we would like to predict).
This further allows us to censor the data manually and examine the effects of such censorship.

We apply manual censorship to the time series of each super-hub in two stages.
In the first stage, for each day $i$ in $N=\{1 \dots 379\}$, we let $\delta_i \in \{0, 1\}$
indicate whether at any moment during $i$ there were no bikes available in the entire super-hub, and define accordingly the set of censored and non-censored observations:
%indicate whether at any moment during $i$ there were no bikes to rent in any hub of the super-hub, and define accordingly the set of censored and non-censored observations:
\begin{align}
    N_{c} &= \{i \in 1..379 : \delta_i = 1\}
    \mathcomma
    \\
    N_{nc} &= \{i \in 1..379 : \delta_i = 0\} = N - N_{c}
    \mathdot
\end{align}
We then fix binary censorship labels as follows: $\lbl_i = 1$ for $i \in N_{c}$ and $\lbl_i = 0$ for $i \in N_{nc}$.
The reason for doing so is that for every day in $N_{c}$, there was a moment with zero bike availability, and so there may have been additional demand, which the service could not satisfy and which was thus not recorded. 

\begin{figure}[tb]
    \centering
    \includegraphics[width=\textwidth]{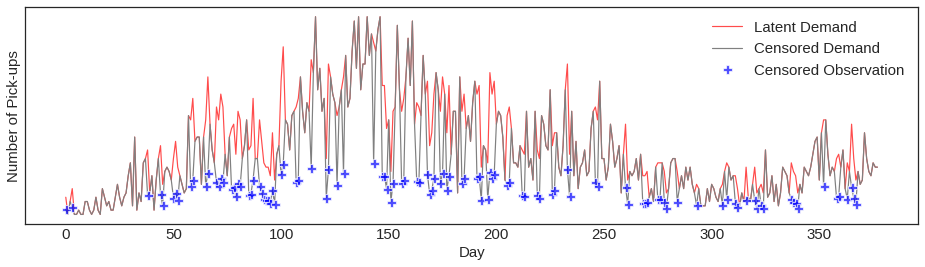}
    \caption{Super-hub 3 data with $50\%$ Censorship Intensity.}
    \label{fig:censoring}
\end{figure}

Having fixed the censorship labels, the second stage of censorship can be executed multiple times for different censorship intensities.
That is, given a censorship intensity $0 \leq c \leq 1$, we censor each observation for which $\lbl_i = 1$ to $(1 - c)$ of its original value.
For example, \fgrref{fig:censoring} shows true demand (red) as well as the result of censoring it to $50\%$ of its original value (grey) for each day in $N_c$ (blue markers).

% The goal of this experiment is, as stated previously, to assess how well different predictive models are able to recover latent bike-sharing demand given a set of censored historical observations. 
% However, in the given experiment, the historical data at our disposal naturally represents only the censored version of the demand, making the evaluation of any given model a non-trivial task.
% Moreover, given the aggregation in super-hubs, the same concept of supply -- and consequently of censorship -- changes. 
% To address this issue, in what follows we will assume that a generic user, in the process of requesting a bike from a specific hub, in case there are no bikes available, is willing to walk to any other hub within the same super-hub in order to rent a bike.
% We thus define a super-hub $S$ as a set of hubs, for which the total availability at time $t$ is
% \begin{equation}
%     a_S(t) = \sum_{s\in S} a_s(t).
% \label{eq:21}
% \end{equation}
% where $a_s(t)$ is the available supply in hub $s$ at time $t$.

\subsubsection{Modeling and Evaluation}

As for previous experiments, we train models on the manually censored demand and then evaluate them on the observations without censorship -- this is the measure of actual interest, as it represents true demand.
Evaluation is done by comparing the posterior mean to the predicted mean through two commonly used measures: coefficient of determination ($R^2$) and Rooted Mean Squared Error (RMSE), as detailed in Appendix \ref{sec:rmse_r2}.

As defined in Section \ref{sec:models}, the three models are \ncgpu, \ncgpa and \cgp.
We equip each model with a kernel that consists of a combination of SE and Periodic kernels on the temporal index feature and a Mat\'ern kernel on weather data.
Namely, the covariance function between the $i$'th and the $j$'th sample is:
\begin{equation}
    k(\vec{x}_i, \vec{x}_j) = k_{SE}(\vec{x}_{i,t}, \vec{x}_{j,t}) + k_{Per}(\vec{x}_{i,t}, \vec{x}_{j,t}) + k_{Mat}(\vec{x}_{i,w}, \vec{x}_{j,w})
    \mathcomma
\label{eq:22}
\end{equation}
where $\vec{x}_{i,t} = \left[ t \right]$ is the temporal index, and $\vec{x}_{i,w}$ consists of weather features as detailed in \tabref{tab:weather}. 
Concretely, we estimate kernel parameters through Type-II Maximum Likelihood using Adam (\cgp) and BFGS (\ncgpu, \ncgpa) to perform gradient ascent.
This choice of optimization routines was the most performing in practice for each respective model.

\begin{table}[tb]
\begin{center}

\begin{tabular}{llll}
\toprule 
Solar Irradiance & Wind & Air & Rain\\
% \hline 
% \hline 
\midrule
Global short wave $[\si{\watt}/\si{\meter}^2]$ & Direction min. $[\si{\degree}]$ & Temperature $[\si{\celsius}]$ & Fall $[\si{\milli\meter}]$\\
Short wave horizontal $[\si{\watt}/\si{\meter}^2]$ & Direction max. $[\si{\degree}]$ & Relative humidity $[\%]$ & Duration $[\si{\second}]$\\
% \hline 
Short wave direct normal $[\si{\watt}/\si{\meter}^2]$ & Direction avg. $[\si{\degree}]$ & Pressure $[\si{\hecto\pascal}]$ & Intensity $[\si{\milli\meter}/\si{\hour}]$\\
% \hline 
Downwelling horizontal long wave $[\si{\watt}/\si{\meter}^2]$ & Speed avg. $[\si{\meter}/\si{\second}]$ &  & \\
% \hline 
 & Speed max. $[\si{\meter}/\si{\second}]$ &  & \\
\bottomrule
\end{tabular}
\caption{Weather features for bike-sharing data, collected in Copenhagen by \cite{dtuclimate}.}
\label{tab:weather}
\end{center}
\end{table}

The main goal of our experiments is to analyze how well these models can recover the \emph{true} demand pattern after being trained on a censored version of the same demand. 
In particular, we are interested in investigating to what degree censored models (\cgp) are able to recover the underlying demand pattern compared to standard regression models (\ncgpu, \ncgpa), and how this comparison evolves under different censorship intensities. 
Our experiments thus employ incremental censorship intensity $c = 0, 0.1, 0.2, \dots, 1.0$, so that $c$ ranges between two extremes: from absence of censoring (\emph{full observability} of the true demand) to complete censoring (no demand observed in historical data).

\subsubsection{Results for Bike-Sharing Experiments} \label{results}

%\begin{figure}[tb]
%\centering
%    \begin{subfigure}[b]{0.6\textwidth}
%        \includegraphics[width=\textwidth]{images/r2_plot.png}
%    \end{subfigure}
%\vspace{3mm}
%    \begin{subfigure}[b]{0.6\textwidth}
%        \includegraphics[width=\textwidth]{images/rmse_plot.png}
%    \end{subfigure}
%\caption{Censoring performance analysis on the entire dataset (left) and only on non-censored observations (right) regarding $R^2$ (top) and RMSE (bottom) for super-hub 1.}
%\label{fig:results_red_superhub}
%\end{figure}

\begin{figure}[h]
    \centering
    \includegraphics[trim={4px 4px 4px 4px},clip,width=\textwidth]{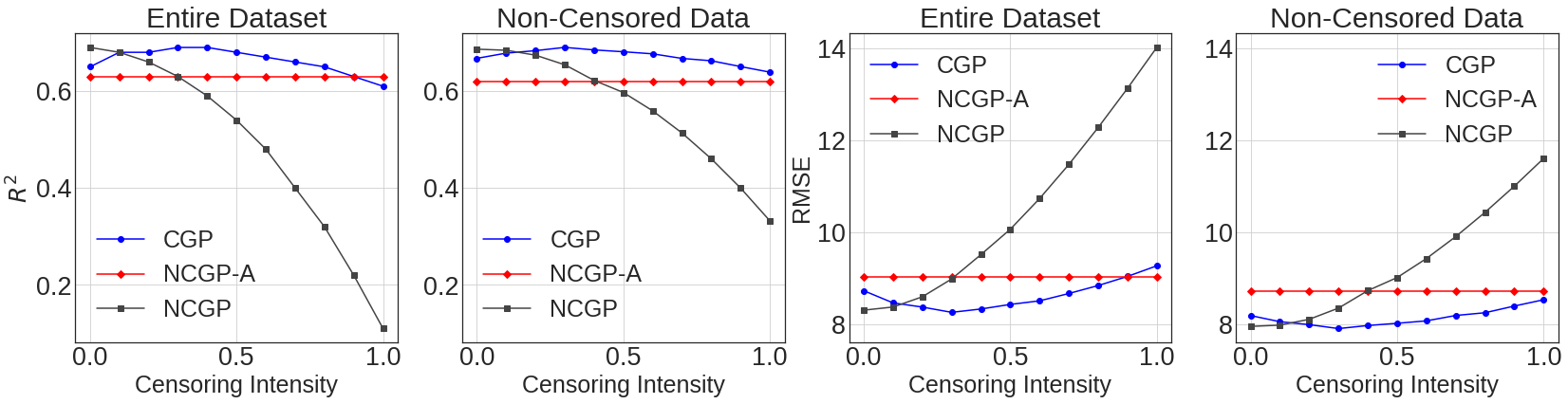}
    \caption{Censoring performance analysis on the entire dataset and only on non-censored observations, regarding $R^2$ and RMSE for super-hub 1.}
    \label{fig:results_red_superhub}
\end{figure}

This section presents results for the predictive models implemented on each of the three time-series with cross-validation.
Tables \ref{tab:superhub1}, \ref{tab:superhub2}, \ref{tab:superhub3} in Appendix \ref{sec:bike_tables} detail the evaluation for each of the three super-hubs.
We now concentrate on the results for super-hub 1, as presented in \fgrvref{fig:results_red_superhub}, because they are representative also of the results for the other two super-hubs.
The plots in \fgrref{fig:results_red_superhub} are a visual representation of \tabref{tab:superhub1} and compare the performances of \ncgpu, \ncgpa and \cgp for different censorship intensities. 
We discern between evaluating model performance on the entire dataset (consisting of both censored and non-censored observations) vs. only on non-censored observations.

First, we compare the models that do not discard of any observations, namely, \cgp and \ncgpu.
Considering that a predictive model is better the more its RMSE is to close $0$ and the more its $R^2$ is close to $1$, the plots show that the two models are comparable under low degree of censoring. However, as the censorship intensifies, \ncgpu becomes strongly biased towards the censored observations, whereas \cgp recovers the underlying demand much more consistently.

Next, we compare between \ncgpa and the \cgp and see that \ncgpa achieves reasonable predictive accuracy, which is still mostly worse than the predictive accuracy of \cgp.
As outlined previously in Sections \ref{sec:synthetic} and \ref{sec:motorcycle}, \ncgpa accuracy depends highly on the extent to which observable data characterizes the full behavior of the latent function (in this case, true demand). 
Here, the percent of points affected by censoring falls between $20\%$ and $40\%$ for all the three super-hubs, so that \ncgpa has acceptable observability over the true demand. 
Even so, \cgp outperforms \ncgpa also on just \emph{non-censored} data; this suggests that using a censored likelihood not only allows models to avoid predictive bias on censored data, but also allows consistent understanding of the data generating process, ultimately leading to increased performance also on observable data. 

In conclusion, and as outlined in previous Sections, the non-parametric nature of Censored GP allows it to effectively exploit the \emph{concept} of censoring, thus preventing censored observations from biasing the entire demand model.
In other words, Censored GP is capable of activating \emph{censoring-awareness} depending on data only.

\subsubsection{Behavior under Zero Censorship}

It is worth emphasizing how Non-Censored GP is a special case of Censored model GP in those cases where all observations are assumed to be non-censored.
This becomes evident when juxtaposing the two likelihood functions:

\begin{align}
    L_C &= 
    \prod_{i \in N_{nc}}
    \left(
    \frac{1}{\sigma} \phi\left(\frac{y_i-f_i}{\sigma}\right)
    \right) 
    \prod_{i \in N_{c}} 
    \left(
    1 - \Phi\left(\frac{y_i-f_i}{\sigma}\right)
    \right)
    \mathcomma
    \\
    L_{NC} &= \prod_{i \in N}
    \left(
    \frac{1}{\sigma} \phi\left(\frac{y_i-f_i}{\sigma}\right)
    \right)
    \mathcomma
\end{align}
which shows that in absence of censoring (i.e., when $N_c=\emptyset$), $L_C \equiv L_{NC}$. 

For $0\%$ censorship, we still see in \fgrref{fig:results_red_superhub} some difference in performance.
To realize why this happens, recall that by our experimental design (Section \ref{sec:bikecens}), censorship labels $\lbl_i$ are pre-determined regardless of censorship intensity, hence some observations are labeled as censored even when censorship intensity is $0$.
Furthermore, the true demand in our experiments originates from historical data, which is itself intrinsically censored. It follows that the models are never exposed to utter absence of censored observations.
Conversely, in a real-world scenario, perfect $0\%$ censoring corresponds to complete absence of censored observations, so that censored and non-censored GP are equivalent.

\subsection{Real-World Dataset 3: Taxi Demand and Supply} \label{sec:taxi}

The previous Section dealt with \emph{long-term} transport demand under \emph{deterministic} censorship per available supply.
In what follows, we complement this with a case study of \emph{short-term} transport demand under \emph{stochastic} censorship per available supply.
To this end, we apply GP modeling to stochastically censored pickups of yellow, hail-based taxis \cite{taxi_data_2010_2013} within an approximately $\SI{1}{\km} \times \SI{1}{\km}$ squared area near LaGuardia airport in New York City.
The true pickup counts, $\vec{y}^* = \left[ \y{*}{1}, \dots, \y{*}{672} \right]$, are aggregated per $\SI{15}{\minute}$ consecutive intervals in the first week of June 2010.
At that time, NYC Taxis were almost entirely of the yellow type and thus accounted for virtually all ride-hailing demand.

\subsubsection{\randdropoff Experiments}

\begin{algorithm}[tb]
\caption{\randdropoff Censorship}
\label{alg:rand_dropoff}
\DontPrintSemicolon
\SetAlgoLined
\SetKwInOut{Input}{input}
\Input{
    $
    0 \leq \gamma \leq 1, 
    0 \leq c \leq 1, 
    \vec{y}^* = \left[\y{*}{1}, \dots, \y{*}{n}\right]^{\mbox{\scriptsize T}} \in \mathbb{R}^n, 
    \vec{d} = \left[d_0, \dots, d_{n - 1}\right]^{\mbox{\scriptsize T}} \in \mathbb{R}^n
    $
    \mathdot
}
\ForEach{$i = 1..n$}{
    Sample $\lbl_i \in \{0, 1\}$ as $\lbl_i \sim \text{Bernoulli}\left(p_{i, \gamma}\right)$, where
    \begin{equation}
    p_{i, \gamma} = 
    \bigg \{
        1 + 
        \exp
        \left(
            \ln \left(\frac{1 - \gamma}{\gamma}\right) 
            - 
            \frac{y_i - d_{i - 1}}{y_i}
        \right)
    \bigg \}^{-1}
    \mathdot
    \end{equation} \\

    $\y{}{i} \gets \y{*}{i} \left( 1 - c\right)^{\lbl_i}$ 
    \mathdot
    \\
}
\Return{$\vec{y}_{} = \left[ \y{}{1}, \dots, \y{}{n} \right]$, $\vec{k} = \left[\lbl_1, \dots, \lbl_n\right]$\mathdot}
\end{algorithm}

\begin{figure}[tb]
    \centering
    \includegraphics[trim={0.9cm 0.9cm 0.7cm 0.7cm}, clip,width=\textwidth]{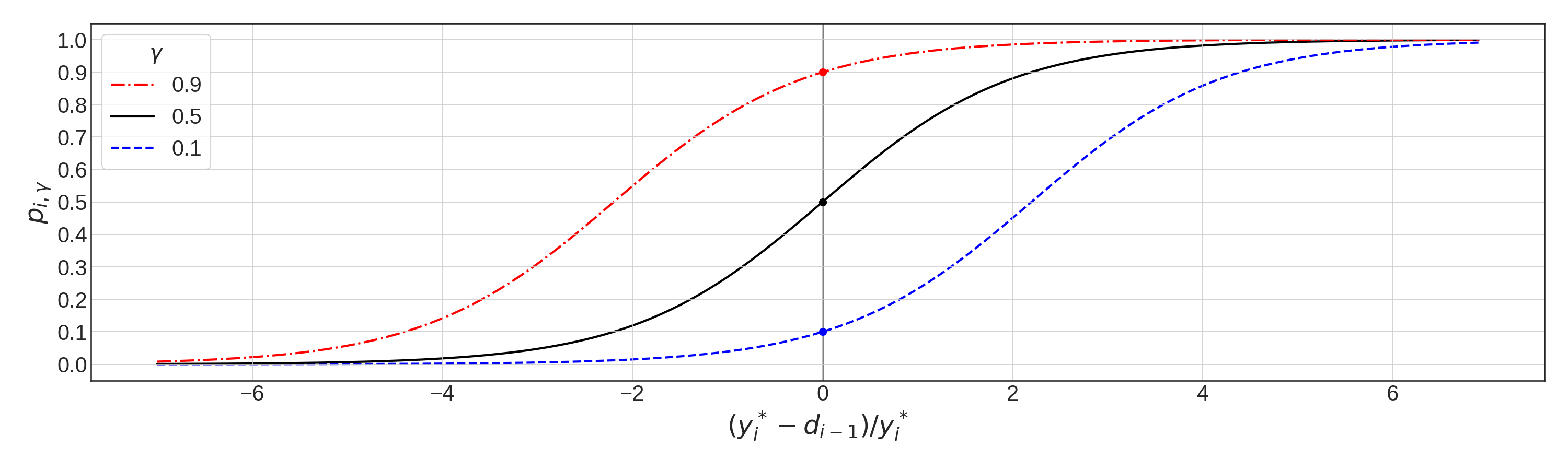}
    \caption{
        In \randdropoff, censorship probability $p_{i, \gamma}$ depends on a zero-intercept parameter $\gamma$ and the difference between pickups $\y{*}{i}$ and dropoffs $d_{i-1}$.
        The higher $\gamma$ is and the higher $\y{*}{i}$ is above $d_{i - 1}$, the more likely will $\y{*}{i}$ be censored.}
    \label{fig:sigmoid_randdropoff}
\end{figure}

\begin{figure}[tb]
    \centering
    \includegraphics[trim={4px 6px 14px 0px},clip,width=\textwidth]{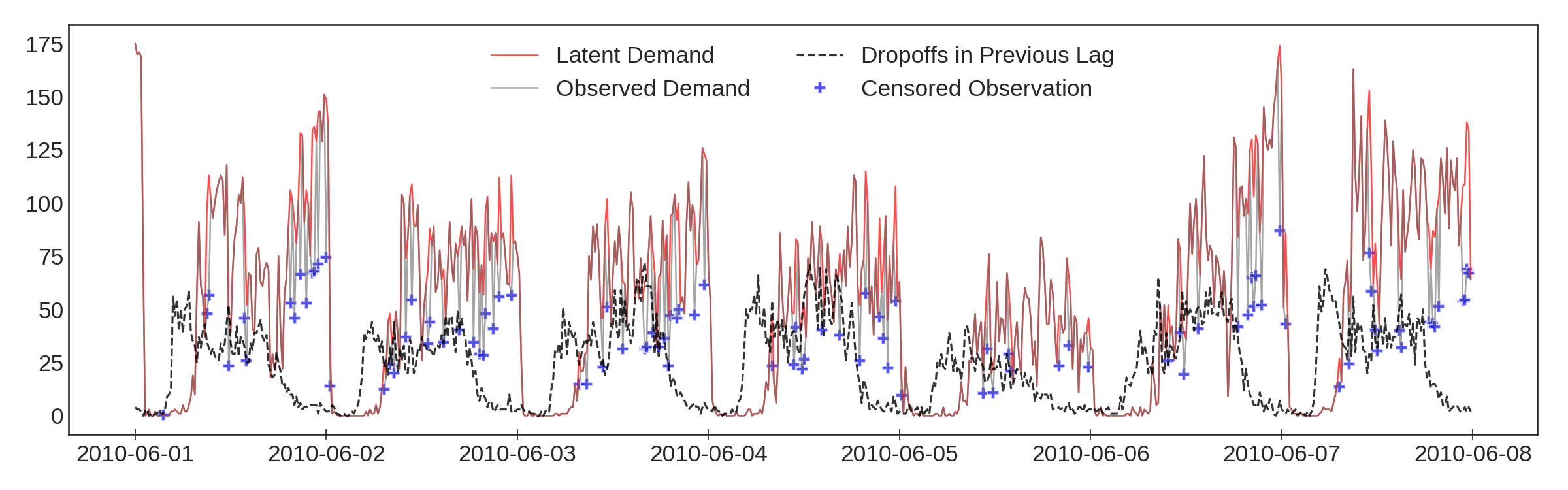}
    \caption{Example of \randdropoff censorship with $\gamma=0.1$ and $c=0.5$.}
    \label{fig:randdropoff_example}
\end{figure}

Taxi demand is influenced by availability of vacant taxis.
To approximate the number of vacant taxis, we use the number of dropoffs observed in the previous lag.
That is, we use $d_{i - 1}$, the observed total dropoffs in time step ${i - 1}$, to stochastically censor $\y{*}{i}$, the true pickups in time step $i$, as follows.
First, we randomly select the censorship label $\lbl_i \in \{0, 1\}$ with probability depending on $\y{*}{i}, d_{i - 1}$, and a parameter $\gamma \in (0, 1)$.
As explained in \fgrref{fig:sigmoid_randdropoff}, approximately $\gamma$ of all $\y{*}{i}$ are thus labeled as censored ($\lbl_i = 1$).
Next, we censor from above each observation for which $\lbl_i = 1$, by setting $\y{}{i}$ to $(1 - c)\y{*}{i}$, where $c \in \left[0, 1\right]$ is a given parameter.
$c$ is thus the percent of pickups that are unobserved and so represents unsatisfied ride-hailing demand.
This two-step procedure, named \randdropoff, is rigorously defined in Algorithm \ref{alg:rand_dropoff}.
\fgrref{fig:randdropoff_example} illustrates an example of \randdropoff for one combination of $\gamma, c$.

We experiment with \randdropoff censorship for $\gamma=0.1, 0.2, 0.3, 0.4$ and $c = 0, 0.1, \dots, 1$.
For these values of $\gamma$, the average percentages of censored observations are, respectively, $13\%, 23\%, 32\%, 40\%$ -- close to the expected values.
We will later see that these values of $\gamma$ suffice for reasoning about the behavior of each model as more observations become censored.
The censorship magnitude increases with $c$, so that $c=0$ and $c=1$ represent limiting cases: when $c$ tends to $0$, censored observations become very close to true values, whereas when $c$ tends to $1$, censored observations are zeroed.

The models are again \ncgpu, \ncgpa and \cgp, and fitting is done via BFGS with at most 1000 iterations, starting from the same initial hyper-parameters.
Because \randdropoff is stochastic, we experiment with each $\gamma, c$ combination independently a total of $30$ times.
In each independent experiment, we fit and evaluate through cross-validation with 21 time-consecutive folds, each consisting of $32$ observations over $8$ hours.
For each $\gamma$ and $c$, we evaluate by comparing $\vec{y}_{}$, the latent count of pickups, to the predicted means over all $30$ experiments, as detailed in Appendix \ref{sec:rmse_r2}.

For all models, the explanatory features are
$
    \mat{X} = \left[\vec{x}_1, \dots, \vec{x}_{672}\right]
$,
where for all time steps $t=1 \dots 672$, $\vec{x}_t$ consists of: $t$, the corresponding hour-of-day in $0 \dots 23$, and the corresponding day-of-week in $0 \dots 6$ (this increases similarity between data points in the same day and improves fit accuracy). 
For \ncgpa and \cgp, $\vec{x}_t$ also contains the binary censorship label $\lbl_t$.
Because weather conditions in the first week of June 2010 were quite stable, they do not serve here as features.
The GP kernel for all three models is thus: 
\begin{equation}
    k\left(\vec{x}, \vec{x}^\prime\right) =
    k_{SE}\left(\vec{x}, \vec{x}^\prime \right)
    + 
    k_{Per}\left(\vec{x}, \vec{x}^\prime \right)
    % =
    % \lambda_{SE}^2 \exp\left( -\frac{\norm{\vec{x} - \vec{x}^\prime}^2}{2\tau_{SE}^2} \right)
    % +
    % \lambda_{Per}^2 \exp\left(-\frac{2 \sin\left(\pi \norm{\vec{x} - \vec{x}^\prime} / \rho\right)}{\tau_{Per}^2} \right)
    \mathdot
    \label{eq:taxikern}
\end{equation}

\subsubsection{Results of \randdropoff Experiments}

We evaluate the models both on the entire dataset and exclusively on non-censored observations.
The results are illustrated in \fgrvref{fig:taxi_performance} and detailed in \tabref{tab:taxi} in Appendix \ref{sec:taxi_tables}.
Next, we analyze the behavior of each model per \fgrref{fig:taxi_performance}.

\begin{figure}[h]
    \centering
    \begin{subfigure}{\textwidth}
        \includegraphics[trim={4px 6px 14px 0px},clip,width=\textwidth]{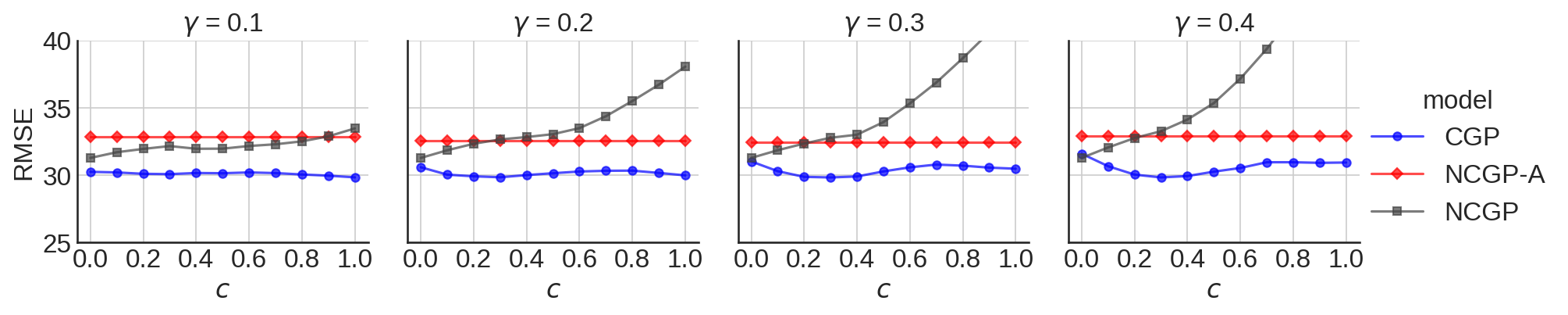}
        \includegraphics[trim={4px 6px 14px 0px},clip,width=\textwidth]{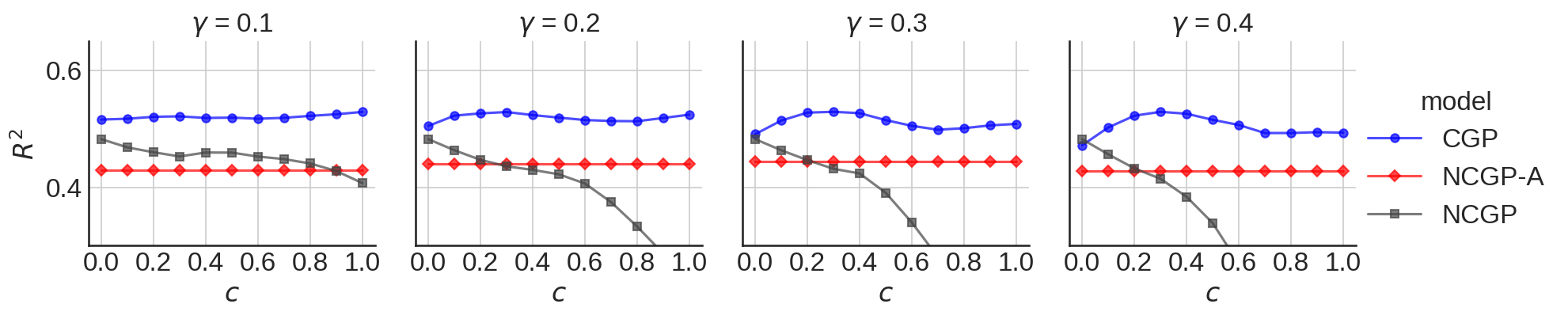}
        \caption{Evaluation over all observations.}
        \label{fig:taxi_eval_all_data}
    \end{subfigure}
    \hfill
    \vspace{0.4cm}
    \begin{subfigure}{\textwidth}
        \includegraphics[width=\textwidth]{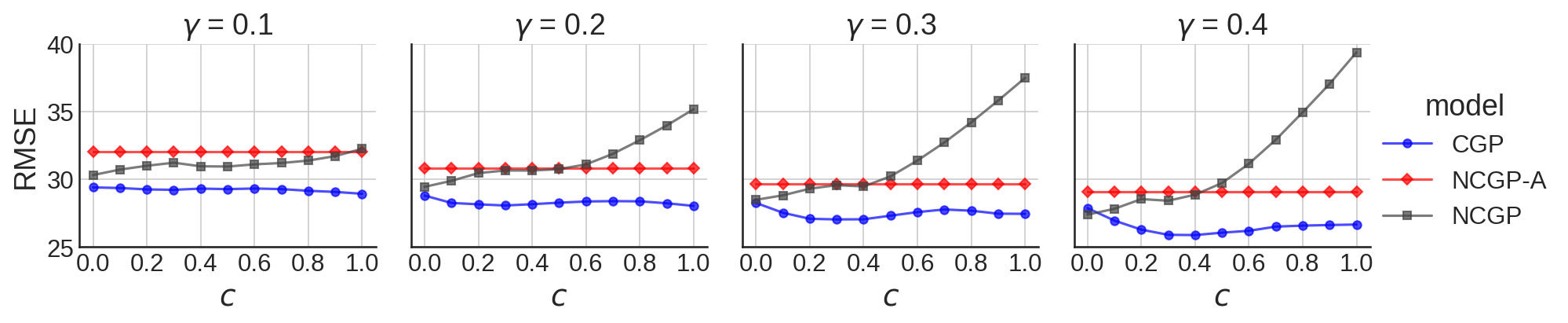}
        \includegraphics[width=\textwidth]{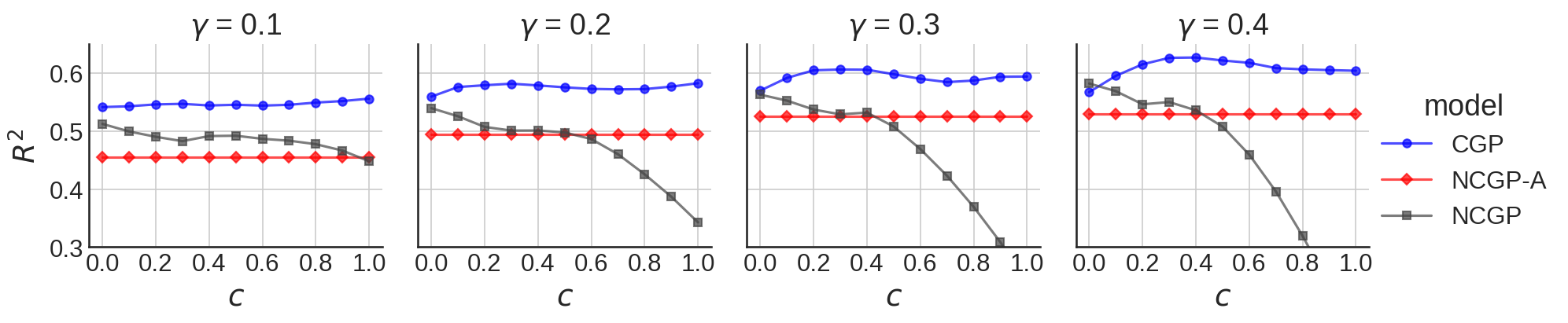}
        \caption{Evaluation over only non-censored observations.}
        \label{fig:taxi_eval_only_non_censored}
    \end{subfigure}
    \caption{Performance of models in \randdropoff experiments. Extreme values of \ncgpu are omitted for easier comparison between \ncgpa and \cgp.}
    \label{fig:taxi_performance}
\end{figure}

As $\gamma$ and $c$ increase, so do the percentage of censored observations and the intensity of censorship.
Consequently, we see in \fgrref{fig:taxi_performance} that \ncgpu deteriorates rapidly as the censored observations draw it downwards, away from the latent pickups.
Conversely, \ncgpa maintains a rather stable performance even as $\gamma$ increases, which suggests that it has enough non-censored points for constructing a stable fit.
Note that as \ncgpa ignores censored observations, its performance does not depend on $c$ and thus traces a horizontal line for each $\gamma$.

For the limiting case of $\gamma=0.4$ and $c=0$, \ncgpu is the best performing mode, slightly better than \cgp.
This, however, may be expected, because labeling many observations as censored without actually censoring them is misleading for any censorship-aware model, including \cgp.
In all other cases, \cgp outperforms \ncgpu and \ncgpa, whether for the entire dataset or for only non-censored observations.
This suggests that \cgp is more reliable not only in coping with censorship, but also in situations where observations are known to accurately reflect the latent ground truth.
Moreover, \cgp performance follows a pattern, which becomes more pronounced as $\gamma$ increases: \cgp starts close to \ncgpu for $c=0$, improves as $c$ increases until about $c=0.5$, then decreases and converges.

Finally, we note that we have also experimented with a similarly sized spatial area in the middle of Manhattan, NYC.
Contrary to the time series used in this Section, the Manhattan cell exhibited a repetitive and regular demand pattern, for which \ncgpa and \cgp performed quite closely.
The noticeably better performance of \cgp in this Section thus suggests that the advantages of censored modeling emerge in more challenging settings --- where the ability to extract meaningful information from censored observations is indeed essential for capturing the underlying demand pattern.
% Opposed to the time-series used in this Section, this cell exhibits a highly repetitive and regular demand pattern where \ncgpa and \cgp performed quite closely. Experiments seem to show how the advantages of censored models emerge especially in more challenging settings, where the ability to extract meaningful information from censored observations can have a significant impact on the correct estimation of the underlying demand pattern.

\section{Summary and Future Work}

Building a model for demand prediction naturally relies on extrapolating knowledge from historical data. 
This is usually done by implementing different types of regression models, to both explain past demand behavior and compute reliable predictions for the future -- a fundamental building block for a great number decision making processes. 
However, we have shown how a reliable predictive model must take into consideration \emph{censoring}, especially in those cases in which demand is implicitly limited by supply. 
More importantly, we stressed the fact that, in the context of shared transport demand modeling, there is a need for models which can deal with censoring in a meaningful way, rather than resorting to different \emph{data cleaning} techniques.

To deal with the censoring problem, we have constructed models that incorporate a \emph{censored} likelihood function within a flexible, non-parametric Gaussian Process (GP).
We compare this model to commonly used GP models, which incorporate a standard Gaussian likelihood, through a series of experiments on synthetic and real-world datasets.
These experiments highlight how standard regression models are prone to return a biased model of demand under data censorship, whereas the proposed Censored GP model yields consistent predictions even under severe censorship.

The experimental results thus confirm the importance of censoring in demand modeling, especially in the transport scenario where demand and supply are naturally interdependent. 
More generally, our results support the idea of building more \emph{knowledgeable} models instead of using case-dependent data cleaning techniques. 
This can be done by feeding the demand models insights on how the demand patterns actually behave, so that the models can adjust automatically to the available data.

For future work, we plan to study settings where censorship labels are partly or even completely unknown.
We also note that shared mobility demand prediction can benefit from utilizing spatio-temporal correlations, as commutes between areas take place regularly and as nearby areas are likely to exhibit similar demand patterns. 
Hence also for future work, we plan to implement models which predict censored demand jointly for multiple areas.
% We note that in this setting of multi-variate modeling, the models can take further advantage of concurrent differences in demand observability, so that an area with fully observed demand could provide useful information for another area where demand is censored.
% By yielding more accurate predictions, such multivariate models of spatio-temporal demand can serve as more reliable tools for decision making. 

\section*{Acknowledgement}

The research leading to these results has received funding from the People Programme (Marie Curie Actions) of the European Union’s Horizon 2020 research and innovation programme under the Marie Sklodowska-Curie Individual Fellowship H2020-MSCA-IF-2016, ID number 745673.

\bibliographystyle{unsrt}  
\bibliography{references}

\newpage
\appendix
\section{Appendix} \label{sec:appendix}

\subsection{Evaluation Measures} \label{sec:rmse_r2}

Models in our experiments are evaluated with the following measures:
\begin{align}
    R^2 &= 1 - \frac{
        \sum_{i=1}^m \left(\hat{\mu}_i - y_i^*\right)^2
        }{
        \sum_{i=1}^m \left(\bar{y}^* - y_i^*\right)^2
        } 
    \mathcomma
    \\
    \mbox{RMSE} &= \sqrt{\frac{1}{n}\sum_{i=1}^m \left(\hat{\mu}_i - y_i^*\right)^2}
    \mathcomma
\end{align}
where $y_1^*, \dots, y_m^*$ are the true demand, $\hat{\mu}_1, \dots, \hat{\mu}_m$ are the corresponding evaluations of the posterior mean, and  
\begin{equation}
\bar{y}^* = \frac{1}{n} \sum_{i=1}^n y_i^*
\end{equation}
is the mean true demand.

\subsection{Experimental Results for \secref{sec:donkey}} \label{sec:bike_tables}

Tables \ref{tab:superhub1}, \ref{tab:superhub2}, \ref{tab:superhub3} hereby provide the results for the experiments of \secref{sec:donkey}, with best values highlighted in bold.

\begin{table}[H]
  \centering
  \caption{Model Performance for Super-hub 1}
  \resizebox{0.85\textwidth}{!}{%
    \begin{tabular}{llccccccccccc}
    \toprule
          \multicolumn{2}{l}{Censorship Intensity:} & 0     & 0.1   & 0.2   & 0.3   & 0.4   & 0.5   & 0.6   & 0.7   & 0.8   & 0.9   & 1 \\
    \midrule
    \multicolumn{13}{c}{Entire Dataset}\\
    \midrule
    \multirow{3}{*}{RMSE} & \ncgpu  & \tabemph{8.31}  & \tabemph{8.38}  & 8.6   & 8.99  & 9.52  & 10.07 & 10.74 & 11.48 & 12.29 & 13.14 & 14.03 \\
          & \ncgpa & 9.03  & 9.03  & 9.03  & 9.03  & 9.03  & 9.03  & 9.03  & 9.03  & 9.03  & \tabemph{9.03}  & \tabemph{9.03} \\
          & \cgp   & 8.73  & 8.47  & \tabemph{8.38}  & \tabemph{8.26}  & \tabemph{8.34}  & \tabemph{8.44}  & \tabemph{8.51}  & \tabemph{8.67}  & \tabemph{8.85}  & 9.04  & 9.28 \\
    \midrule
    \multirow{3}{*}{$R^2$} & \ncgpu  & \tabemph{0.69}  & \tabemph{0.68}  & 0.66  & 0.63  & 0.59  & 0.54  & 0.48  & 0.40  & 0.32  & 0.22  & 0.11 \\
          & \ncgpa & 0.63  & 0.63  & 0.63  & 0.63  & 0.63  & 0.63  & 0.63  & 0.63  & 0.63  & \tabemph{0.63}  & 0.63 \\
          & \cgp   & 0.65  & \tabemph{0.68}  & \tabemph{0.68}  & \tabemph{0.69}  & \tabemph{0.69}  & \tabemph{0.68}  & \tabemph{0.67}  & \tabemph{0.66}  & \tabemph{0.65}  & \tabemph{0.63}  & 0.61 \\
    \midrule
    \multicolumn{13}{c}{Non-Censored Data}\\
    \midrule
    \multirow{3}{*}{RMSE} & \ncgpu  & \tabemph{7.96}  & \tabemph{7.99}  & 8.11  & 8.36  & 8.74  & 9.02  & 9.43  & 9.91  & 10.44 & 11.01 & 11.61 \\
          & \ncgpa & 8.73  & 8.73  & 8.73  & 8.73  & 8.73  & 8.73  & 8.73  & 8.73  & 8.73  & 8.73  & 8.73 \\
          & \cgp   & 8.19  & 8.06  & \tabemph{8.00}  & \tabemph{7.91}  & \tabemph{7.98}  & \tabemph{8.02}  & \tabemph{8.08}  & \tabemph{8.20}  & \tabemph{8.25}  & \tabemph{8.40}  & \tabemph{8.54} \\
    \midrule
    \multirow{3}{*}{$R^2$} & \ncgpu  & \tabemph{0.69}  & \tabemph{0.68}  & 0.67  & 0.65  & 0.62  & 0.60  & 0.56  & 0.51  & 0.46  & 0.40  & 0.33 \\
          & \ncgpa & 0.62  & 0.62  & 0.62  & 0.62  & 0.62  & 0.62  & 0.62  & 0.62  & 0.62  & 0.62  & 0.62 \\
          & \cgp   & 0.67  & \tabemph{0.68}  & \tabemph{0.68}  & \tabemph{0.69}  & \tabemph{0.68}  & \tabemph{0.68}  & \tabemph{0.68}  & \tabemph{0.67}  & \tabemph{0.66}  & \tabemph{0.65}  & \tabemph{0.64} \\
    \bottomrule
    \end{tabular}%
    }
  \label{tab:superhub1}%
\end{table}%
% \vspace{-0.6cm}
\begin{table}[H]
  \centering
  \caption{Model Performance for Super-hub 2}
  \resizebox{0.85\textwidth}{!}{%
    \begin{tabular}{llccccccccccc}
    \toprule
          \multicolumn{2}{l}{Censorship Intensity:} & 0     & 0.1   & 0.2   & 0.3   & 0.4   & 0.5   & 0.6   & 0.7   & 0.8   & 0.9   & 1 \\
    \midrule
    \multicolumn{13}{c}{Entire Dataset}\\
    \midrule
    \multirow{3}{*}{RMSE} & \ncgpu  & \tabemph{6.17}  & \tabemph{6.22}  & \tabemph{6.37}  & 6.63  & 7.00     & 7.42  & 7.91  & 8.46  & 9.05  & 9.67  & 10.33 \\
          & \ncgpa & 7.28  & 7.28  & 7.28  & 7.28  & 7.28  & 7.28  & 7.28  & 7.28  & 7.28  & 7.28  & \tabemph{7.28} \\
          & \cgp   & 6.74  & 6.57  & 6.44  & \tabemph{6.43}  & \tabemph{6.52}  & \tabemph{6.59}  & \tabemph{6.72}  & \tabemph{6.90}  & \tabemph{7.05}  & \tabemph{7.22}  & 7.40 \\
    \midrule
    \multirow{3}{*}{$R^2$} & \ncgpu  & \tabemph{0.65}  & \tabemph{0.65}  & \tabemph{0.63}  & 0.60  & 0.56  & 0.50  & 0.43  & 0.35  & 0.26  & 0.15  & 0.03 \\
          & \ncgpa & 0.52  & 0.52  & 0.52  & 0.52  & 0.52  & 0.52  & 0.52  & 0.52  & 0.52  & 0.52  & \tabemph{0.52} \\
          & \cgp   & 0.59  & 0.61  & 0.62  & \tabemph{0.62}  & \tabemph{0.61}  & \tabemph{0.61}  & \tabemph{0.59}  & \tabemph{0.57}  & \tabemph{0.55}  & \tabemph{0.53}  & 0.50 \\
    \midrule
    \multicolumn{13}{c}{Non-Censored Data}\\
    \midrule
    \multirow{3}{*}{RMSE} & \ncgpu  & \tabemph{5.85}  & \tabemph{5.88}  & \tabemph{5.99}  & 6.16  & 6.39  & 6.65  & 6.96  & 7.31  & 7.70  & 8.12  & 8.56 \\
          & \ncgpa & 6.72  & 6.72  & 6.72  & 6.72  & 6.72  & 6.72  & 6.72  & 6.72  & 6.72  & 6.72  & 6.72 \\
          & \cgp   & 6.21  & 6.11  & 6.01  & \tabemph{6.01}  & \tabemph{6.05}  & \tabemph{6.09}  & \tabemph{6.13}  & \tabemph{6.28}  & \tabemph{6.42}  & \tabemph{6.51}  & \tabemph{6.68} \\
    \midrule
    \multirow{3}{*}{$R^2$} & \ncgpu  & \tabemph{0.71}  & \tabemph{0.71}  & \tabemph{0.70}  & 0.68  & 0.66  & 0.63  & 0.59  & 0.55  & 0.50  & 0.45  & 0.38 \\
          & \ncgpa & 0.62  & 0.62  & 0.62  & 0.62  & 0.62  & 0.62  & 0.62  & 0.62  & 0.62  & 0.62  & \tabemph{0.62} \\
          & \cgp   & 0.68  & 0.69  & \tabemph{0.70}  & \tabemph{0.70}  & \tabemph{0.69}  & \tabemph{0.69}  & \tabemph{0.68}  & \tabemph{0.67}  & \tabemph{0.65}  & \tabemph{0.64}  & \tabemph{0.62} \\
    \bottomrule
    \end{tabular}%
    }
  \label{tab:superhub2}%
\end{table}%
% \vspace{-0.6cm}
\begin{table}[H] % Table generated by Excel2LaTeX from sheet 'TS3'
  \centering
  \caption{Model Performance for Super-hub 3}
  \resizebox{0.85\textwidth}{!}{%
    \begin{tabular}{llccccccccccc}
    \toprule
          \multicolumn{2}{l}{Censorship Intensity:} & 0     & 0.1   & 0.2   & 0.3   & 0.4   & 0.5   & 0.6   & 0.7   & 0.8   & 0.9   & 1 \\
    \midrule
    \multicolumn{13}{c}{Entire Dataset} \\
    \midrule
    \multirow{3}{*}{RMSE} & \ncgpu  & \tabemph{7.27}  & 7.32  & 7.45  & 7.62  & 7.87  & 8.19  & 8.56  & 8.99  & 9.46  & 9.96  & 10.5 \\
          & \ncgpa & 7.35  & 7.35  & 7.35  & 7.35  & 7.35  & 7.35  & 7.35  & \tabemph{7.35}  & \tabemph{7.35}  & \tabemph{7.35}  & \tabemph{7.35} \\
          & \cgp   & 7.42  & \tabemph{7.26}  & \tabemph{7.20}  & \tabemph{7.17}  & \tabemph{7.19}  & \tabemph{7.25}  & \tabemph{7.29}  & 7.37  & 7.42  & 7.50  & 7.65 \\
\midrule
\multirow{3}{*}{$R^2$} & \ncgpu  & \tabemph{0.54}  & 0.53  & 0.51  & 0.49  & 0.46  & 0.41  & 0.36  & 0.29  & 0.22  & 0.13  & 0.03 \\
          & \ncgpa & 0.53  & 0.53  & 0.53  & 0.53  & 0.53  & 0.53  & \tabemph{0.53}  & \tabemph{0.53}  & \tabemph{0.53}  & \tabemph{0.53}  & \tabemph{0.53} \\
          & \cgp   & 0.52  & \tabemph{0.54}  & \tabemph{0.55}  & \tabemph{0.55}  & \tabemph{0.55}  & \tabemph{0.54}  & \tabemph{0.53}  & 0.52  & 0.52  & 0.51  & 0.49 \\
    \midrule
    \multicolumn{13}{c}{Non-Censored Data} \\
    \midrule
    \multirow{3}{*}{RMSE} & \ncgpu  & \tabemph{6.99}  & 7.02  & 7.10  & 7.20  & 7.36  & 7.58  & 7.85  & 8.16  & 8.50  & 8.88  & 9.28 \\
          & \ncgpa & 7.02  & 7.02  & 7.02  & 7.02  & 7.02  & 7.02  & 7.02  & \tabemph{7.02}  & \tabemph{7.02}  & \tabemph{7.02}  & \tabemph{7.02} \\
          & \cgp   & 7.12  & \tabemph{7.01}  & \tabemph{6.97}  & \tabemph{6.92}  & \tabemph{6.96}  & \tabemph{6.98}  & \tabemph{6.99}  & 7.05  & 7.04  & 7.09  & 7.22 \\
    \midrule
    \multirow{3}{*}{$R^2$} & \ncgpu  & \tabemph{0.55}  & 0.55  & 0.54  & 0.53  & 0.50  & 0.47  & 0.44  & 0.39  & 0.34  & 0.28  & 0.21 \\
          & \ncgpa & 0.55  & \tabemph{0.55}  & 0.55  & 0.55  & 0.55  & \tabemph{0.55}  & \tabemph{0.55}  & \tabemph{0.55}  & \tabemph{0.55}  & \tabemph{0.55}  & \tabemph{0.55} \\
          & \cgp   & 0.54  & \tabemph{0.55}  & \tabemph{0.56}  & \tabemph{0.56}  & \tabemph{0.56}  & \tabemph{0.55}  & \tabemph{0.55}  & 0.54  & 0.55  & 0.54  & 0.52 \\
    \bottomrule
    \end{tabular}%
    }
  \label{tab:superhub3}%
\end{table}%

\subsection{Experimental Results for \secref{sec:taxi}} \label{sec:taxi_tables}

\tabref{tab:taxi} hereby details the experimental results for \secref{sec:taxi}, with best values highlighted in bold.

\begin{table}[H]
\centering
\caption{Model Performance in \randdropoff Experiments}
\resizebox{1\textwidth}{!}{%
\begin{tabular}{
    r
    r
    |
    p{\dimexpr 0.085\linewidth-2\tabcolsep}
    p{\dimexpr 0.105\linewidth-2\tabcolsep}
    p{\dimexpr 0.085\linewidth-2\tabcolsep}
    |
    p{\dimexpr 0.09\linewidth-2\tabcolsep}
    p{\dimexpr 0.11\linewidth-2\tabcolsep}
    p{\dimexpr 0.075\linewidth-2\tabcolsep}
    |
    p{\dimexpr 0.09\linewidth-2\tabcolsep}
    p{\dimexpr 0.11\linewidth-2\tabcolsep}
    p{\dimexpr 0.085\linewidth-2\tabcolsep}
    |
    p{\dimexpr 0.09\linewidth-2\tabcolsep}
    p{\dimexpr 0.11\linewidth-2\tabcolsep}
    p{\dimexpr 0.07\linewidth-2\tabcolsep}
}
\toprule
 & & \multicolumn{6}{c}{Entire Dataset} & \multicolumn{6}{c}{Only Non-Censored} \\
 & & \multicolumn{3}{c}{RMSE} & \multicolumn{3}{c}{$R^2$} & \multicolumn{3}{c}{RMSE} & \multicolumn{3}{c}{$R^2$} \\
$\gamma$ & $c$ & \ncgpu & \ncgpa & \cgp & \ncgpu & \ncgpa & \cgp & \ncgpu & \ncgpa & \cgp & \ncgpu & \ncgpa & \cgp \\
\toprule
	.1 & 0 & $31.295$ & $32.867$ & $\tabemph{30.261}$  & $.482$ & $.429$ & $\tabemph{.516}$  & $30.319$ & $32.036$ & $\tabemph{29.405}$  & $.512$ & $.455$ & $\tabemph{.541}$  \\
	 & .1 & $31.717$ & $32.867$ & $\tabemph{30.216}$  & $.468$ & $.429$ & $\tabemph{.517}$  & $30.721$ & $32.036$ & $\tabemph{29.360}$  & $.499$ & $.455$ & $\tabemph{.543}$  \\
	 & .2 & $31.964$ & $32.867$ & $\tabemph{30.112}$  & $.460$ & $.429$ & $\tabemph{.521}$  & $31.002$ & $32.036$ & $\tabemph{29.257}$  & $.490$ & $.455$ & $\tabemph{.546}$  \\
	 & .3 & $32.174$ & $32.867$ & $\tabemph{30.089}$  & $.453$ & $.429$ & $\tabemph{.521}$  & $31.228$ & $32.036$ & $\tabemph{29.225}$  & $.482$ & $.455$ & $\tabemph{.547}$  \\
	 & .4 & $31.975$ & $32.867$ & $\tabemph{30.174}$  & $.460$ & $.429$ & $\tabemph{.519}$  & $30.963$ & $32.036$ & $\tabemph{29.316}$  & $.491$ & $.455$ & $\tabemph{.544}$  \\
	 & .5 & $31.982$ & $32.867$ & $\tabemph{30.156}$  & $.459$ & $.429$ & $\tabemph{.519}$  & $30.945$ & $32.036$ & $\tabemph{29.278}$  & $.492$ & $.455$ & $\tabemph{.545}$  \\
	 & .6 & $32.180$ & $32.867$ & $\tabemph{30.217}$  & $.453$ & $.429$ & $\tabemph{.517}$  & $31.108$ & $32.036$ & $\tabemph{29.323}$  & $.486$ & $.455$ & $\tabemph{.544}$  \\
	 & .7 & $32.308$ & $32.867$ & $\tabemph{30.167}$  & $.448$ & $.429$ & $\tabemph{.519}$  & $31.204$ & $32.036$ & $\tabemph{29.279}$  & $.483$ & $.455$ & $\tabemph{.545}$  \\
	 & .8 & $32.525$ & $32.867$ & $\tabemph{30.063}$  & $.441$ & $.429$ & $\tabemph{.522}$  & $31.385$ & $32.036$ & $\tabemph{29.155}$  & $.477$ & $.455$ & $\tabemph{.549}$  \\
	 & .9 & $32.898$ & $32.867$ & $\tabemph{29.976}$  & $.428$ & $.429$ & $\tabemph{.525}$  & $31.714$ & $32.036$ & $\tabemph{29.071}$  & $.466$ & $.455$ & $\tabemph{.551}$  \\
	 & 1 & $33.487$ & $32.867$ & $\tabemph{29.849}$  & $.407$ & $.429$ & $\tabemph{.529}$  & $32.256$ & $32.036$ & $\tabemph{28.935}$  & $.448$ & $.455$ & $\tabemph{.556}$  \\
\midrule
	.2 & 0 & $31.294$ & $32.555$ & $\tabemph{30.601}$  & $.482$ & $.440$ & $\tabemph{.505}$  & $29.446$ & $30.838$ & $\tabemph{28.803}$  & $.539$ & $.495$ & $\tabemph{.559}$  \\
	 & .1 & $31.862$ & $32.555$ & $\tabemph{30.050}$  & $.463$ & $.440$ & $\tabemph{.523}$  & $29.889$ & $30.838$ & $\tabemph{28.258}$  & $.525$ & $.495$ & $\tabemph{.576}$  \\
	 & .2 & $32.347$ & $32.555$ & $\tabemph{29.928}$  & $.447$ & $.440$ & $\tabemph{.527}$  & $30.460$ & $30.838$ & $\tabemph{28.155}$  & $.507$ & $.495$ & $\tabemph{.579}$  \\
	 & .3 & $32.668$ & $32.555$ & $\tabemph{29.855}$  & $.436$ & $.440$ & $\tabemph{.529}$  & $30.654$ & $30.838$ & $\tabemph{28.068}$  & $.501$ & $.495$ & $\tabemph{.581}$  \\
	 & .4 & $32.845$ & $32.555$ & $\tabemph{30.019}$  & $.430$ & $.440$ & $\tabemph{.524}$  & $30.651$ & $30.838$ & $\tabemph{28.166}$  & $.501$ & $.495$ & $\tabemph{.578}$  \\
	 & .5 & $33.060$ & $32.555$ & $\tabemph{30.156}$  & $.422$ & $.440$ & $\tabemph{.519}$  & $30.753$ & $30.838$ & $\tabemph{28.276}$  & $.497$ & $.495$ & $\tabemph{.575}$  \\
	 & .6 & $33.509$ & $32.555$ & $\tabemph{30.284}$  & $.406$ & $.440$ & $\tabemph{.515}$  & $31.101$ & $30.838$ & $\tabemph{28.365}$  & $.486$ & $.495$ & $\tabemph{.572}$  \\
	 & .7 & $34.394$ & $32.555$ & $\tabemph{30.337}$  & $.375$ & $.440$ & $\tabemph{.514}$  & $31.882$ & $30.838$ & $\tabemph{28.390}$  & $.460$ & $.495$ & $\tabemph{.572}$  \\
	 & .8 & $35.526$ & $32.555$ & $\tabemph{30.350}$  & $.333$ & $.440$ & $\tabemph{.513}$  & $32.887$ & $30.838$ & $\tabemph{28.377}$  & $.425$ & $.495$ & $\tabemph{.572}$  \\
	 & .9 & $36.736$ & $32.555$ & $\tabemph{30.178}$  & $.287$ & $.440$ & $\tabemph{.519}$  & $33.974$ & $30.838$ & $\tabemph{28.235}$  & $.387$ & $.495$ & $\tabemph{.576}$  \\
	 & 1 & $38.077$ & $32.555$ & $\tabemph{29.998}$  & $.234$ & $.440$ & $\tabemph{.524}$  & $35.175$ & $30.838$ & $\tabemph{28.038}$  & $.343$ & $.495$ & $\tabemph{.582}$  \\
\midrule
	.3 & 0 & $31.294$ & $32.444$ & $\tabemph{31.032}$  & $.482$ & $.444$ & $\tabemph{.491}$  & $28.489$ & $29.673$ & $\tabemph{28.259}$  & $.563$ & $.525$ & $\tabemph{.570}$  \\
	 & .1 & $31.864$ & $32.444$ & $\tabemph{30.310}$  & $.463$ & $.444$ & $\tabemph{.514}$  & $28.820$ & $29.673$ & $\tabemph{27.532}$  & $.552$ & $.525$ & $\tabemph{.591}$  \\
	 & .2 & $32.360$ & $32.444$ & $\tabemph{29.887}$  & $.446$ & $.444$ & $\tabemph{.528}$  & $29.311$ & $29.673$ & $\tabemph{27.087}$  & $.537$ & $.525$ & $\tabemph{.604}$  \\
	 & .3 & $32.793$ & $32.444$ & $\tabemph{29.843}$  & $.432$ & $.444$ & $\tabemph{.529}$  & $29.571$ & $29.673$ & $\tabemph{27.041}$  & $.529$ & $.525$ & $\tabemph{.606}$  \\
	 & .4 & $33.019$ & $32.444$ & $\tabemph{29.917}$  & $.424$ & $.444$ & $\tabemph{.527}$  & $29.476$ & $29.673$ & $\tabemph{27.051}$  & $.532$ & $.525$ & $\tabemph{.606}$  \\
	 & .5 & $33.956$ & $32.444$ & $\tabemph{30.291}$  & $.391$ & $.444$ & $\tabemph{.515}$  & $30.220$ & $29.673$ & $\tabemph{27.318}$  & $.508$ & $.525$ & $\tabemph{.598}$  \\
	 & .6 & $35.344$ & $32.444$ & $\tabemph{30.594}$  & $.340$ & $.444$ & $\tabemph{.505}$  & $31.390$ & $29.673$ & $\tabemph{27.582}$  & $.469$ & $.525$ & $\tabemph{.590}$  \\
	 & .7 & $36.898$ & $32.444$ & $\tabemph{30.800}$  & $.280$ & $.444$ & $\tabemph{.499}$  & $32.719$ & $29.673$ & $\tabemph{27.775}$  & $.423$ & $.525$ & $\tabemph{.584}$  \\
	 & .8 & $38.704$ & $32.444$ & $\tabemph{30.719}$  & $.208$ & $.444$ & $\tabemph{.501}$  & $34.206$ & $29.673$ & $\tabemph{27.685}$  & $.369$ & $.525$ & $\tabemph{.587}$  \\
	 & .9 & $40.624$ & $32.444$ & $\tabemph{30.572}$  & $.128$ & $.444$ & $\tabemph{.506}$  & $35.814$ & $29.673$ & $\tabemph{27.466}$  & $.309$ & $.525$ & $\tabemph{.593}$  \\
	 & 1 & $42.594$ & $32.444$ & $\tabemph{30.492}$  & $.041$ & $.444$ & $\tabemph{.509}$  & $37.487$ & $29.673$ & $\tabemph{27.453}$  & $.242$ & $.525$ & $\tabemph{.594}$  \\
\midrule
	.4 & 0 & $\tabemph{31.297}$ & $32.914$ & $31.607$  & $\tabemph{.482}$ & $.427$ & $.472$  & $\tabemph{27.395}$ & $29.079$ & $27.873$  & $\tabemph{.582}$ & $.529$ & $.567$  \\
	 & .1 & $32.070$ & $32.914$ & $\tabemph{30.679}$  & $.456$ & $.427$ & $\tabemph{.502}$  & $27.824$ & $29.079$ & $\tabemph{26.956}$  & $.568$ & $.529$ & $\tabemph{.595}$  \\
	 & .2 & $32.770$ & $32.914$ & $\tabemph{30.054}$  & $.432$ & $.427$ & $\tabemph{.523}$  & $28.541$ & $29.079$ & $\tabemph{26.287}$  & $.546$ & $.529$ & $\tabemph{.615}$  \\
	 & .3 & $33.279$ & $32.914$ & $\tabemph{29.844}$  & $.415$ & $.427$ & $\tabemph{.529}$  & $28.420$ & $29.079$ & $\tabemph{25.904}$  & $.550$ & $.529$ & $\tabemph{.626}$  \\
	 & .4 & $34.145$ & $32.914$ & $\tabemph{29.946}$  & $.384$ & $.427$ & $\tabemph{.526}$  & $28.868$ & $29.079$ & $\tabemph{25.886}$  & $.536$ & $.529$ & $\tabemph{.627}$  \\
	 & .5 & $35.369$ & $32.914$ & $\tabemph{30.274}$  & $.339$ & $.427$ & $\tabemph{.516}$  & $29.726$ & $29.079$ & $\tabemph{26.063}$  & $.507$ & $.529$ & $\tabemph{.621}$  \\
	 & .6 & $37.184$ & $32.914$ & $\tabemph{30.547}$  & $.269$ & $.427$ & $\tabemph{.507}$  & $31.149$ & $29.079$ & $\tabemph{26.200}$  & $.459$ & $.529$ & $\tabemph{.617}$  \\
	 & .7 & $39.396$ & $32.914$ & $\tabemph{30.973}$  & $.180$ & $.427$ & $\tabemph{.493}$  & $32.926$ & $29.079$ & $\tabemph{26.510}$  & $.396$ & $.529$ & $\tabemph{.608}$  \\
	 & .8 & $41.949$ & $32.914$ & $\tabemph{30.973}$  & $.070$ & $.427$ & $\tabemph{.493}$  & $34.950$ & $29.079$ & $\tabemph{26.583}$  & $.319$ & $.529$ & $\tabemph{.606}$  \\
	 & .9 & $44.679$ & $32.914$ & $\tabemph{30.929}$  & $-.055$ & $.427$ & $\tabemph{.494}$  & $37.051$ & $29.079$ & $\tabemph{26.627}$  & $.235$ & $.529$ & $\tabemph{.605}$  \\
	 & 1 & $47.582$ & $32.914$ & $\tabemph{30.953}$  & $-0.197$ & $.427$ & $\tabemph{.494}$  & $39.378$ & $29.079$ & $\tabemph{26.664}$  & $.136$ & $.529$ & $\tabemph{.604}$  \\
	 \bottomrule
\end{tabular}
} % End of resizebox
\label{tab:taxi}
\end{table}

\end{document}